\documentclass[journal, letterpaper]{IEEEtran}

\usepackage[T1]{fontenc}
\usepackage{textcomp}
\usepackage{graphicx}
\usepackage{url}        
\usepackage{amsmath}    
\usepackage{authblk}

\usepackage{textgreek}	% Greek to me, dawg
\usepackage{listings}
\usepackage{longtable}
\usepackage{array}
\usepackage{hyperref}
\usepackage[
    style=numeric,
    sorting=none,
    defernumbers=false,
    citestyle=numeric,
    bibstyle=numeric
]{biblatex}
\DeclareFieldFormat{doi}{}
\DeclareFieldFormat{url}{}
\newcommand{\mytableref}[1]{Table \ref{#1}}

\title{Robust Monocular Visual Odometry using Curriculum Learning}
\author{Assaf Lahiany\textsuperscript{1}, Oren Gal\textsuperscript{1}}

\vspace{20pt}  % Added space between author and first affiliation
\affil{\textsuperscript{1}Swarm \& AI Lab (SAIL)\\Hatter Department of Marine Technologies\\Leon H. Charney School of Marine Sciences\\University of Haifa}

\begin{document}

\maketitle

\begin{abstract}
	Curriculum Learning (CL), drawing inspiration from natural learning patterns observed in humans and animals, employs a systematic approach of gradually introducing increasingly complex training data during model development. Our work applies innovative CL methodologies to address the challenging geometric problem of monocular Visual Odometry (VO) estimation, which is essential for robot navigation in constrained environments. The primary objective of our research is to push the boundaries of current state-of-the-art (SOTA) benchmarks in monocular VO by investigating various curriculum learning strategies. We enhance the end-to-end Deep-Patch-Visual Odometry (DPVO) framework through the integration of novel CL approaches, with the goal of developing more resilient models capable of maintaining high performance across challenging environments and complex motion scenarios. Our research encompasses several distinctive CL strategies. We develop methods to evaluate sample difficulty based on trajectory motion characteristics, implement sophisticated adaptive scheduling through Self-Paced weighted loss mechanisms, and utilize reinforcement learning agents for dynamic adjustment of training emphasis. Through comprehensive evaluation on the diverse synthetic TartanAir dataset and complex real-world benchmarks such as EuRoC and TUM-RGBD, our Curriculum Learning-based Deep-Patch-Visual Odometry (CL-DPVO) demonstrates superior performance compared to existing SOTA methods, including both feature-based and learning-based VO approaches. The results validate the effectiveness of integrating curriculum learning principles into visual odometry systems.
\end{abstract}

\section{Introduction}
	
	 Visual Odometry (VO) is a crucial technique in robotics and computer vision that estimates an agent's egomotion, specifically, its position and orientation, based on visual input. While VO has shown promising results in controlled environments, its application in critical real-world scenarios, especially when sensors like GPS, LiDAR, and Inertial Measurement Units (IMUs) cannot be used, presents significant challenges that can compromise accuracy or lead to system failure. The performance of VO systems is particularly susceptible to dynamic motion patterns. High-frequency movements, abrupt camera tilts, and rapid maneuvers can introduce noise and discontinuities in the visual stream, complicating the extraction of reliable motion estimates. These challenges are further exacerbated in less-than-ideal, and often adverse, environmental conditions. A robust VO model must demonstrate resilience across a spectrum of visual contexts. Low-light scenarios, for instance, reduce the visibility of salient features necessary for accurate tracking. Motion blur, resulting from relative movement between the camera and the environment, introduces additional complexity to feature detection and matching algorithms. These multifaceted challenges underscore the need for advanced VO algorithms and training techniques capable of maintaining accuracy and reliability across a wide range of operational conditions. As such, addressing these issues remains a critical focus in the ongoing development of robust visual odometry systems for autonomous navigation and localization.

    To address these challenges, we propose novel curriculum learning strategies integrated into the Deep-Patch-Visual-Odometry (DPVO) framework. Our approach prioritizes intelligent training methodologies over complex multi-modal architectures, aiming to enhance model robustness and reduce training resources while preserving inference computational efficiency.

\section{Background}

\subsection{Visual Odometry: Foundational Concept and Challenges}

    Visual odometry (VO) has been a focal point of research in robotics and computer vision, with monocular VO gaining particular attention due to its cost-effectiveness and simplicity. Traditional monocular VO methods primarily relied on hand-crafted features and geometric techniques, as demonstrated by Nistér et al. (2004) and Scaramuzza and Fraundorfer (2011). While effective in controlled environments, these methods often face challenges in complex real-world scenarios due to scale ambiguity and sensitivity to environmental changes. The emergence of deep learning has revolutionized VO research, including monocular approaches. Kendall et al. (2015) introduced PoseNet \cite{kendall2015posenet}, marking one of the first deep learning methods for camera relocalization, which laid the groundwork for end-to-end learning of pose estimation directly from images. Building on this foundation, Wang et al. (2017) developed DeepVO \cite{wang2017deepvo}, showcasing the potential of recurrent neural networks to capture temporal dependencies in monocular VO tasks.

    Recent advancements have focused on enhancing the robustness of deep learning-based monocular VO systems. Zhan et al. (2019) introduced Unsupervised VO with Geometric Constraints (UnDeepVO) \cite{8461251}, which leverages unsupervised learning to address the limitations of supervised methods in real-world settings. Saputra et al. (2019) \cite{saputra2019learning} proposed a novel approach that combines geometric and learning-based techniques to improve performance in challenging environments. Additionally, researchers like Zachary et al. (2024) \cite{lipson2024deep} and Klenk et al. (2024) \cite{klenk2024deep} have employed deep patch selection mechanisms to further enhance monocular model accuracy and efficiency, including an advanced event-based variation.

    However, augmenting a model's proficiency in adapting to varied motion dynamics and visual degradation remains a significant challenge. Several methodologies have been proposed to mitigate input distortions and variability, yet each harbors inherent limitations:

	\noindent
    \textbf{Preprocessing Enhancement:} Utilizing techniques such as image deblurring \cite{schuler2015learning} and super-resolution \cite{9025375} before model inference. Although beneficial in certain contexts, this strategy results in information loss stemming from the enhancement models' presupposed priors on "clean" data statistics.

	\noindent
    \textbf{Single Model with Diverse Training:} This involves training one model across a broad spectrum of input qualities and distortions. This method often necessitates extensive datasets and more sophisticated models for effective generalization \cite{yang2018quality}, \cite{dodge2018quality}, \cite{dodge2016understanding}, \cite{liu2017quality}.

	\noindent
    \textbf{Ensemble Methods:} Training multiple models, each tailored to specific distortion ranges \cite{campos2021orb}, \cite{fraundorfer2012visual}. While this method can be effective, it does not facilitate information exchange among models, thus limiting its ability to generalize across all input quality variations.

	\noindent
    \textbf{Data \& Sensor Fusion:} By integrating traditional camera imagery with event-based camera data, learning-based and model-based approaches have shown promise in improving accuracy under difficult conditions. Nonetheless, the management and integration of these diverse data sources significantly increase computational demands and system complexity \cite{hidalgo2022event}, \cite{clark2017vinet}, \cite{yuan2022robust}.

\subsection{Curriculum learning}

    Curriculum learning (CL), introduced by Bengio et al. in 2009 \cite{weinshall2018curriculum}, offers a potential solution to these challenges. This approach involves designing a "training curriculum" that progressively introduces more difficult examples during the training process. Recent applications of curriculum learning in computer vision have shown promising results: Jiang et al. \cite{jiang2015self} presented a curriculum-based CNN for scene classification, where the training curriculum was based on image difficulty defined by the source of the image. In the domain of image segmentation, Wei et al. \cite{wei2016stc} proposed a curriculum learning approach where an initial model is trained on simple images using saliency maps, followed by the progressive inclusion of more complex samples. Weinshall et al. \cite{weinshall2018curriculum} investigated the robustness of curriculum learning across various computer vision tasks, highlighting its superiority in convergence compared to standard training methods.

\subsection{Curriculum Learning in Visual Odometry}

    A critical aspect of curriculum learning is the requirement for explicit labels of task complexity for each training instance. In the context of visual odometry, this can be achieved by applying synthetic augmentations with controlled parameters (e.g., noise levels, blur, resolution degradation) to clean inputs and the use of diverse dynamic motion scenarios (e.g., maximum translation and rotation speed in recorded trajectories). Hacohen and Weinshall (2019) introduced a method for automatically determining the difficulty of training examples by combining transfer learning from teacher network, which could be adapted for VO tasks \cite{hacohen2019power}. Another notable work includes Saputra et al. (2019) \cite{saputra2019learning}, which presented a novel CL strategy for learning the geometry of monocular VO by gradually making the learning objective more difficult during training using geometry-aware objective function.

\section{Methodology}

    We propose a comprehensive curriculum learning framework for training Deep Visual Odometry systems that adaptively controls the learning progression across multiple components of the visual odometry task, enhancing model performance in real-world scenarios with different motion complexities and environmental conditions. Our approach implements both trajectory based difficulty assignment and dynamic progression strategies to optimize the training trajectory. The curriculum learning system manages three critical aspects of the visual odometry problem: optical flow estimation, pose prediction, and rotation estimation. Each component's difficulty is independently controlled through interpolation weights between initial (simpler) and final (more challenging) configurations. The framework incorporates three methodological approaches: 1) a trajectory-Based approach where difficulties are precalculated based on camera motion characteristics and scene complexity metrics, 2) a Self-Paced learning strategy that dynamically adjusts the curriculum based on current loss values, and 3) an adaptive Deep Deterministic Policy Gradient (DDPG) based scheduler that learns to optimize the curriculum through reinforcement learning. While the first approach rely on predefined, per-trajectory progression scheme, the latter two dynamically adapt the difficulty levels in response to the model's performance, with Self-Paced learning using direct loss feedback and DDPG learning a more complex policy through experience.

\subsection{Datasets}

    We utilize the TartanAir \cite{wang2020tartanair} dataset to both train and assess our curriculum learning methodologies. TartanAir, recognized as a leading benchmark in monocular Visual Odometry (VO) since its inception in 2020, offers significant advantages due to its synthetic origin. Leveraging the Unreal Engine, TartanAir provides environments with high-fidelity realism, allowing for meticulous control over the scenarios used in training. The dataset's sequences exhibit a wide range of motion profiles and environmental complexities, ideal for curriculum learning where the gradual increase in task difficulty is key. Such variability is essential for crafting resilient monocular VO algorithms, addressing challenges like scale ambiguity and drift inherent to single-camera systems. Moreover, the synthetic dataset's capacity to deliver accurate ground truth data for camera poses and optical flow facilitates a rigorous evaluation of monocular VO algorithms, devoid of the typical noise found in real-world datasets.

    To bridge the gap between simulation and reality, we further evaluate our curriculum learning (CL) models using the TUM-RGBD \cite{sturm2012benchmark} and EuRoC \cite{burri2016euroc} benchmarks. The TUM-RGBD dataset consists of sequences captured in various office-like settings with RGB-D sensors, providing depth information alongside color images. This dataset is particularly useful for testing algorithms in scenarios that involve complex object textures and changing lighting conditions, which are typical in indoor environments. On the other hand, the EuRoC dataset, collected from a micro aerial vehicle, includes both indoor and outdoor sequences, offering a different set of challenges like high-speed motion, motion blur, and varying illumination, which are crucial for testing the robustness and generalization capabilities of VO systems in dynamic real-world conditions.
	By employing these datasets, we aim to validate the transferability of our CL approach from synthetic to real-world scenarios.

	Additionally, we complement our real-world validation by evaluating our CL models on the ICL-NUIM \cite{handa2014benchmark} dataset, which offers synthetic sequences with realistic camera noise models, enabling assessment of model robustness to sensor noise. We benchmark against state-of-the-art monocular VO systems across all datasets to demonstrate practical applicability.

\subsection{Baseline Model}

    As our baseline model architecture, we use the Deep Patch Visual Odometry (DPVO) model, introduced in \cite{lipson2024deep}. DPVO represents a state-of-the-art (SOTA) approach to monocular visual odometry, demonstrating competitive performance across standard benchmarks through a patch-based deep learning framework. A key strength of DPVO lies in its end-to-end trainable nature and its inference computational efficiency allowing high FPS with minimal memory requirements. The learning-based end-to-end nature allows the model to learn more robust patch representations and matching strategies directly from data, while the patch-based approach provides better handling of local image structures. Upon its publication, DPVO demonstrated superior performance compared to existing monocular VO methods across standard evaluation metrics, including those utilizing comprehensive SLAM frameworks like DROID-SLAM \cite{teed2021droid}. Subsequently, DPVO has established itself as a fundamental baseline for numerous learning-based architectural enhancements, notably Deep-Event-Visual-Odometry (DEVO) \cite{klenk2024deep}, which leverages simulated event data to enhance system robustness. Given its proven capabilities, DPVO serves as an ideal baseline for investigating the impact of curriculum learning strategies, where we maintain its architecture and hyper-parameters while modifying the training procedure and objective function through our proposed curriculum learning framework.

	\vspace{0.5em}
	\noindent
    \textbf{DPVO Loss Supervision:} At a high level, The DPVO approach works similarly to a classical VO system: it samples a set of patches for each video frame, estimates the 2D motion (optical flow) of each patch against each of its connected frames in patch graph, and solves for depth and camera poses that are consistent with the 2D motions. This approach differs from a classical system in that these steps are done through a recurrent neural network (update operator) and a differentiable optimization layer. DPVO apply supervision to poses and induced optical flow (i.e. trajectory updates), supervising each intermediate output of the update operator and detach the poses and patches from the gradient tape prior to each update. In its core, the DPVO define two types of supervisions:
	\textit{Pose Supervision}. By scaling the predicted trajectory to match the ground truth using the Umeyama alignment algorithm \cite{umeyama1991least}, every pair of poses (i, j), is supervise on the error:
    \begin{equation}
	\sum_{(i,j)\ i \neq j}^{}\left\| {Log}_{SE(3)}\lbrack\left( G_{i}^{- 1}G_{j} \right)^{- 1}(T_{i}^{- 1}T_{j})\rbrack \right\|\tag{1}
    \label{eq:poseSupervision}
	\end{equation}
	where G is the ground truth and T are the predicted poses. 
	\textit{Flow Supervision}. employs supervision based on the difference between predicted and ground truth optical flow fields within a ±2-frame temporal window. Both supervisions are incorporated into the overall loss through weighted summation:
    \begin{equation}
	\mathcal{L}_{total} = 10\mathcal{L}_{pose} + 0.1\mathcal{L}_{flow}\tag{2}
    \label{eq:totalLoss}
	\end{equation}
	with weights empirically determined in \cite{lipson2024deep} to optimize performance metrics while maintaining appropriate scaling between components.

\subsection{Hierarchical Curriculum-Learning Loss Structure}

    Our curriculum learning framework modifies the DPVO training objective by introducing weighted loss components that adapt throughout the training process. The total loss is structured as a nested weighted combination of flow, translation, and rotation components.
    \begin{equation}
	\mathcal{L}_{pose} = \mathcal{(L}_{translation} + w_{r}\mathcal{L}_{rotation})\tag{3}
	\label{eq:CLposeLoss}
    \end{equation}
	\vspace{-17pt}
    \begin{equation}
	\mathcal{L}_{total} = w_{f}s_{f}\mathcal{L}_{flow} + w_{p}s_{p}\mathcal{L}_{pose}\tag{4}
	\label{eq:CLtotalLoss}
    \end{equation}
    This hierarchical weighting scheme implements curriculum learning at multiple levels:
	
	\noindent
    \textbf{Base Weights}~(\(s_{f},\ s_{p}\)): Fixed weights that balance the fundamental trade-off between flow and pose estimation tasks. Ensuring correct magnitude scaling of the flow and pose losses. During our experiments we use \(s_{f} = 0.1\), \(s_{p} = 10\) as in (\ref{eq:totalLoss}).
    
	\noindent
	\textbf{Curriculum Weights}~(\(w_{f}\),\(w_{p}\),\(w_{r}\)): Dynamic weights controlled by the curriculum learning scheduler that adjust the learning emphasis throughout training. \(w_{f}\) controls the importance of optical flow estimation, \(w_{p}\) modulates the overall pose learning (affecting both translation and rotation) and \(w_{r}\) specifically adjusts the rotation component within pose estimation. The hierarchical weighting scheme in (\ref{eq:CLposeLoss}) and (\ref{eq:CLtotalLoss}) enables independent control over the learning progression of each component while preserving their natural relationships. The rotation weight \(w_{r}\) operates within the broader pose estimation weight \(w_{p}\), allowing fine-grained control over rotation learning while maintaining the overall pose learning trajectory. Translation learning is implicitly controlled through the pose weight without requiring a separate translation weight, as it represents the fundamental aspect of pose estimation. This simplified structure reduces complexity by respecting the natural hierarchy of the visual odometry task, where translation serves as the base component of pose estimation, while rotation requires additional fine-tuning through its dedicated weight.

\subsection{Curriculum-Learning Strategies}

\subsubsection{Trajectory-Based}
	Our trajectory-Based curriculum learning approach begins by preprocessing the dataset to quantify motion complexity. For each sequence, we analyze frame-to-frame pose differences to compute maximum translational and rotational magnitudes. These values are normalized and combined into a weighted average difficulty score.

\begin{figure}[ht!]
	    \begin{center}
		\includegraphics[scale=.27]{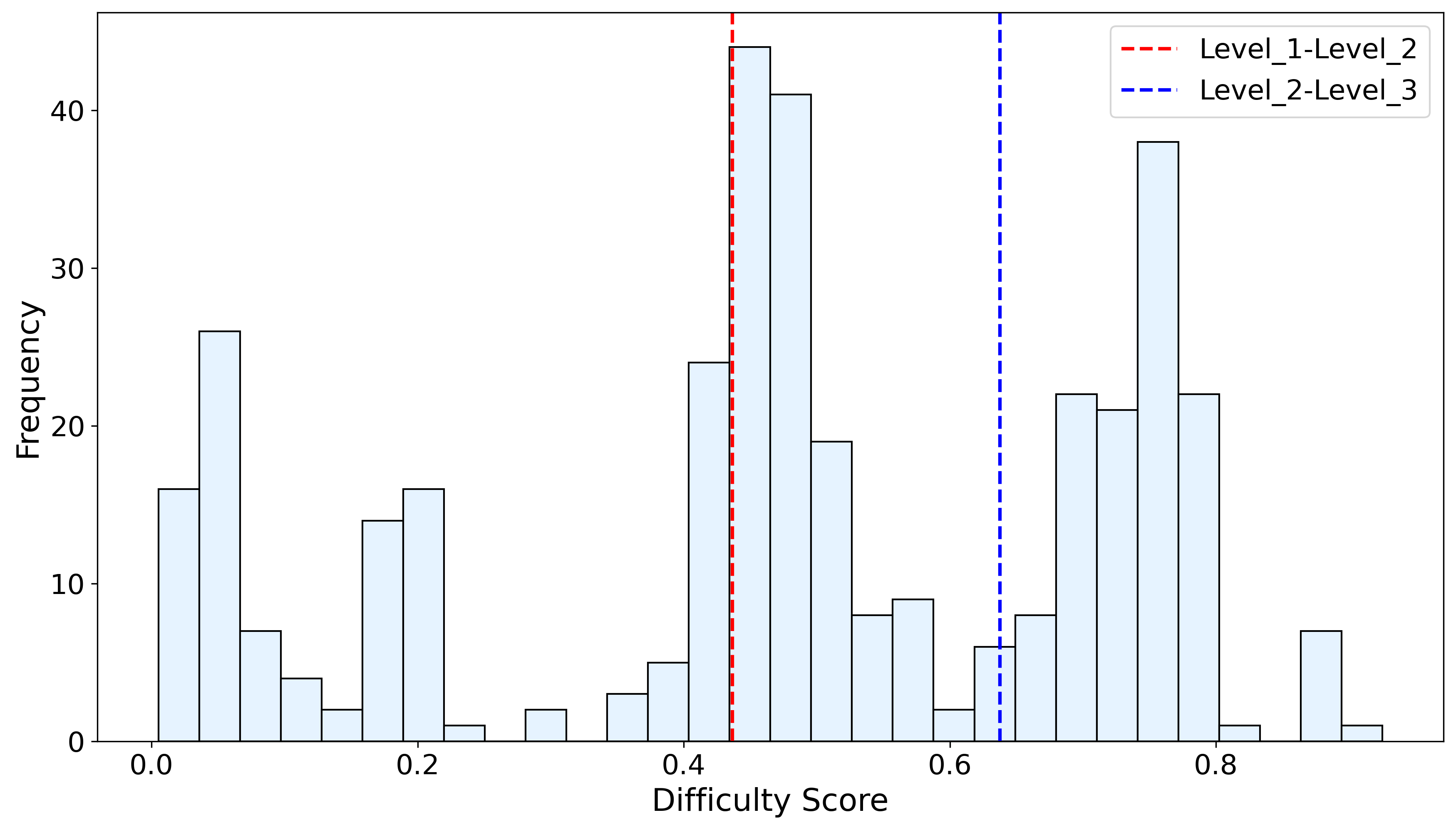}
    	\caption{Distribution of difficulty scores across the TartanAir training dataset, with three difficulty levels, with dashed lines indicating difficulty thresholds at 0.44 and 0.64.}
    	\label{fig:difficultyDistribution}
    	\end{center}
	\end{figure}

	The histogram in \autoref{fig:difficultyDistribution} shows the TartanAir dataset distribution of difficulty scores across three training levels. Sequences with lower scores represent easier training examples, characterized by smaller maximumtranslation and rotation magnitudes. Two vertical dashed lines at scores 0.44 and 0.64 divide the dataset into equal-sized difficulty groups. As shown in \mytableref{tab:diffScoresThresholds}, these groups correspond to easy (level 1), medium (level 2), and hard (level 3) difficulties. The distribution exhibits peaks at scores 0.1, 0.45, and 0.7, demonstrating a balanced spread of difficulty levels across the training data.
	
    \mytableref{tab:diffScoresThresholds} shows the normalized difficulty score ranges for each level.
    \begin{table}[!hbt]

		\begin{center}
		\renewcommand{\arraystretch}{1.2}
		\caption{Difficulty levels complexity corresponding the normalized score range. The difficulty thresholds ensure even number of elements in each difficulty level.}
		\label{tab:diffScoresThresholds}
		\begin{tabular}{|c|c|c|}

			\hline
			\hline
			Level & Difficulty & Normalized Score \\
			
            \hline
			1 & Easy & 0-0.44 \\
            2 & Medium & 0.44-0.64 \\
            3 & Hard & 0.64-1 \\
			\hline
		\end{tabular}
		\end{center}
		
	\end{table}

    Our motion complexity scoring approach extends beyond \cite{wang2020tartanair} by incorporating all six degrees of freedom (DoF) across the full spectrum of difficulty levels. This comprehensive evaluation provides a more complete assessment of motion complexity. During model training, the curriculum progresses through these levels sequentially. Initial training phase focuses exclusively on Level 1 trajectories to establish basic trajectory estimation capabilities, then gradually introduces Level 2 trajectories once performance stabilizes, and finally incorporates Level 3 trajectories to achieve robustness to challenging motion patterns. This three-tier approach introduce a variant to the classic curriculum learning approach in \cite{10.1145/1553374.1553380}. It provides clear transition points in the curriculum while maintaining a manageable complexity progression, allowing the model to build competency in handling increasingly challenging scenarios.

	Throughout the Trajectory-Based training phases, we employ the original DPVO objective function (\ref{eq:totalLoss}). This represents a special case of our curriculum-learning hierarchical loss structure where base weights in (\ref{eq:CLtotalLoss}) are set to 0.1 and 10, while curriculum weights in (\ref{eq:CLposeLoss}) and (\ref{eq:CLtotalLoss}) are fixed at 1.
	\vspace{0.5em}
\subsubsection{Self-Paced Progression}

    Our Self-Paced learning strategy implements a dynamic curriculum that adapts based on the model's current performance, measured through the loss magnitude. This method calculates an adaptive progress factor \(\phi\) that is dependent on the current total loss \(\mathcal{L}_{i}\) and a Self-Paced factor \(\lambda\) which controls the sensitivity to loss changes and act as an adaptive regularization mechanism incorporated into the total loss. This exponential formulation (\ref{eq:selfPacedFactor}) creates an inverse relationship between loss and progress: during training phases where the loss is high (mainly in the early training phase), the exponential term approaches zero, keeping weights closer to their initial values; as the loss decreases, the exponential term approaches one, allowing weights to progress toward their final values. During our experiments, we bound our weights range by setting our initial and final weights to \(w_{0} = 0.1\) and \(w_{F} = 1\) respectively for all components.
    \begin{equation}
    w_{i}^{f,p,r} = w_{0} + (w_{F} - w_{0})\phi(\mathcal{L}_{i})\tag{5}
	\label{eq:selfPacedWeights}
	\end{equation}
	\vspace{-17pt}
	\begin{equation}
    \phi(\mathcal{L}_{i}) = e^{- \lambda\mathcal{L}_{i}}\tag{6}
    \label{eq:selfPacedFactor}
	\end{equation}
	Where \emph{i} is the training step index. The negativeexponential function was chosen for several key properties that make it particularly suitable for curriculum learning. It naturally bounds the adaptive progress between 0 and 1 (assuming loss is always positive), ensuring stable interpolation between initial and final weights (\ref{eq:selfPacedWeights}). The function provides a smooth, continuous progression that avoids abrupt changes in difficulty, while its sensitivity can be precisely controlled through the Self-Paced factor \(\lambda\). The tuning of \(\lambda\) is crucial: a larger \(\lambda\) makes the system more sensitive to loss changes, causing faster adaptation but potentially leading to unstable progression, while a smaller \(\lambda\) provides more gradual changes but might slow down learning. In our implementation, we empirically determined \(\lambda\) through a series of experiments, starting with a moderate value (\(\lambda\)=0.1) and adjusting based on observed learning stability and progression speed. The optimal \(\lambda\) value typically depends on the scale of the loss values and the desired progression rate, requiring careful validation to balance between responsive adaptation and stable learning. This approach provides automatic adaptation to the model's learning pace without requiring predefined training durations or manual scheduling, though the effectiveness depends on careful tuning of the Self-Paced factor \(\lambda\) to achieve optimal training progression.

	\vspace{0.5em}
\subsubsection{Adaptive Learning}

Our adaptive curriculum strategy employs Deep Deterministic Policy Gradient (DDPG) agents to dynamically control component-specific difficulty weights. These Reinforcement-Learning (RL) agents independently regulate the weights for flow estimation, pose prediction, and rotation components. DDPG is specifically designed for continuous action spaces, making it naturally suited for our need to output continuous weight values between 0 and 1 for our curriculum progression. Its actor-critic architecture provides stable learning in continuous domains, where the critic helps reduce the variance of policy updates while the actor learns a deterministic policy. Another advantage of our curriculum learning approach lies in DDPG's off-policy nature, which enables efficient learning through experience replay memory. This allows the agent to learn from past experiences and maintain training stability while adapting to different curriculum phases. Each component weight-DDPG agent is formulated as follows:
    \begin{equation}
    w_{i}^{f,p,r} = w_{0} + (w_{F} - w_{0})a_{i}\tag{7}
	\label{eq:adaptiveWeights}
	\end{equation}
	\vspace{-17pt}
	\begin{equation}
    s_{i} = [p_{i},\ \mathcal{L}_{i}^{f,p,r}]\tag{8}
	\label{eq:adaptiveState}
	\end{equation}
	\vspace{-17pt}
	\begin{equation}
    a_{i} = \mu_{k}(s_{i}) + \mathcal{N}_{i}\tag{9}
	\label{eq:adaptiveAction}
	\end{equation}
	\vspace{-17pt}
	\begin{equation}
    r_{i} = - |\mathcal{L}_{i}^{f,p,r}|\tag{10}
	\label{eq:adaptiveReward}
	\end{equation}

    Each agent observes a~state \(s_{i}\ \)comprising the normalized training progress \(p_{i} = i/N\) (\emph{N} is the bounded estimated total number of training steps) and the current component-specific loss value \(\mathcal{L}_{i}\) (\ref{eq:adaptiveState}). The progress factor \(p_{i}\) is normalized using a predefined total number of steps \emph{N}, chosen primarily for implementation simplicity. However, this can be replaced by more dynamic approaches such as adaptive step counting based on validation performance or hybrid approaches that combine step counts with performance metrics.
    
    Based on state \(s_{i}\), the agent actor's network \emph{µ} outputs an action value between 0 and 1, which is used to interpolate between initial and final weights, \(\left( w_{0},\ w_{F} \right)\) for that weight component (\ref{eq:adaptiveWeights}). The learning process uses the raw loss values \(\mathcal{L}_{i}\ \)as immediate feedback for each specific agent. The reward signal \(r_{i}\) is designed as the negative absolute loss, encouraging the agents to find weight configurations that minimize their respective loss components (\ref{eq:adaptiveReward}). To maintain~exploration throughout training, the agent employs an adaptive~noise mechanism \(\mathcal{N}_{i}\ \) that scales based on the current action values, ensuring appropriate exploration-exploitation balance. This noise is added to the output of the DDPG's actor network \emph{µ} to produce action \(a_{i}\) which is the corresponding CL weight (\ref{eq:adaptiveAction}). Network updates for the agent\textquotesingle s actor-critic architecture are~performed at fixed intervals during the global VO~training process. At each update step, multiple training~iterations are executed on randomly sampled batches from each agent's dedicated experience replay buffer, ensuring thorough learning~from diverse historical experiences. This approach~allows each component\textquotesingle s difficulty to be independently~adjusted based on its specific~learning progress. The DDPG agents are trained concurrently with the VO model, learning to optimize the curriculum through direct~experience with the training dynamics. By leveraging the continuous~action~space of DDPG, these agents~can make fine-grained adjustments to the model\textquotesingle s~weights, potentially leading to more efficient and~effective learning.

\section{Experiments}

    Our experimental evaluation of our curriculum learning strategies employed the TartanAir dataset for validation and testing, enabling us to assess performance against the ECCV 2022 SLAM competition metrics. Our experimental setup preserved DPVO's default architectural hyper-parameters, enabling direct performance comparisons and isolating the impact of our methodology enhancements. As in the default configuration in \cite{lipson2024deep}, we use 96 image patches for feature extraction and a 10-frame sliding window for trajectory optimization. Following the original DPVO evaluation protocol in \cite{lipson2024deep}, we prioritize average trajectory error (ATE) and area under curve (AUC) as our primary validation and testing metrics, as it provides a more realistic indication of real-world performance. Our training infrastructure utilized an NVIDIA DGX-1 computing node equipped with 8XV100 GPUs, facilitating parallel processing and rapid experimental validation across our multiple methodological variants. 
    
\subsection{Training}

    Our curriculum learning (CL) implementation is integrated into the DPVO training pipeline through a dedicated CL scheduler. This scheduler dynamically manages curriculum weights throughout the training process, with updates occurring at each step during loss computation. Before implementing our curriculum learning strategies, we first established a reliable baseline by reproducing the original DPVO results on both TartanAir's validation and test splits. This required adjusting the learning rate to properly support our multi-GPU training setup, ensuring our modifications wouldn't compromise the model's baseline performance. To prevent overfitting and ensure optimal model selection, we implemented an early stopping mechanism that monitors both the AUC and ATE metrics on the validation set. This dual-metric approach provides a more robust criterion for determining when to halt training, as it considers both the model's overall trajectory accuracy and its performance distribution across different scenarios.
    
\begin{figure}[ht!]
        \begin{center}
    	\includegraphics[scale=.11]{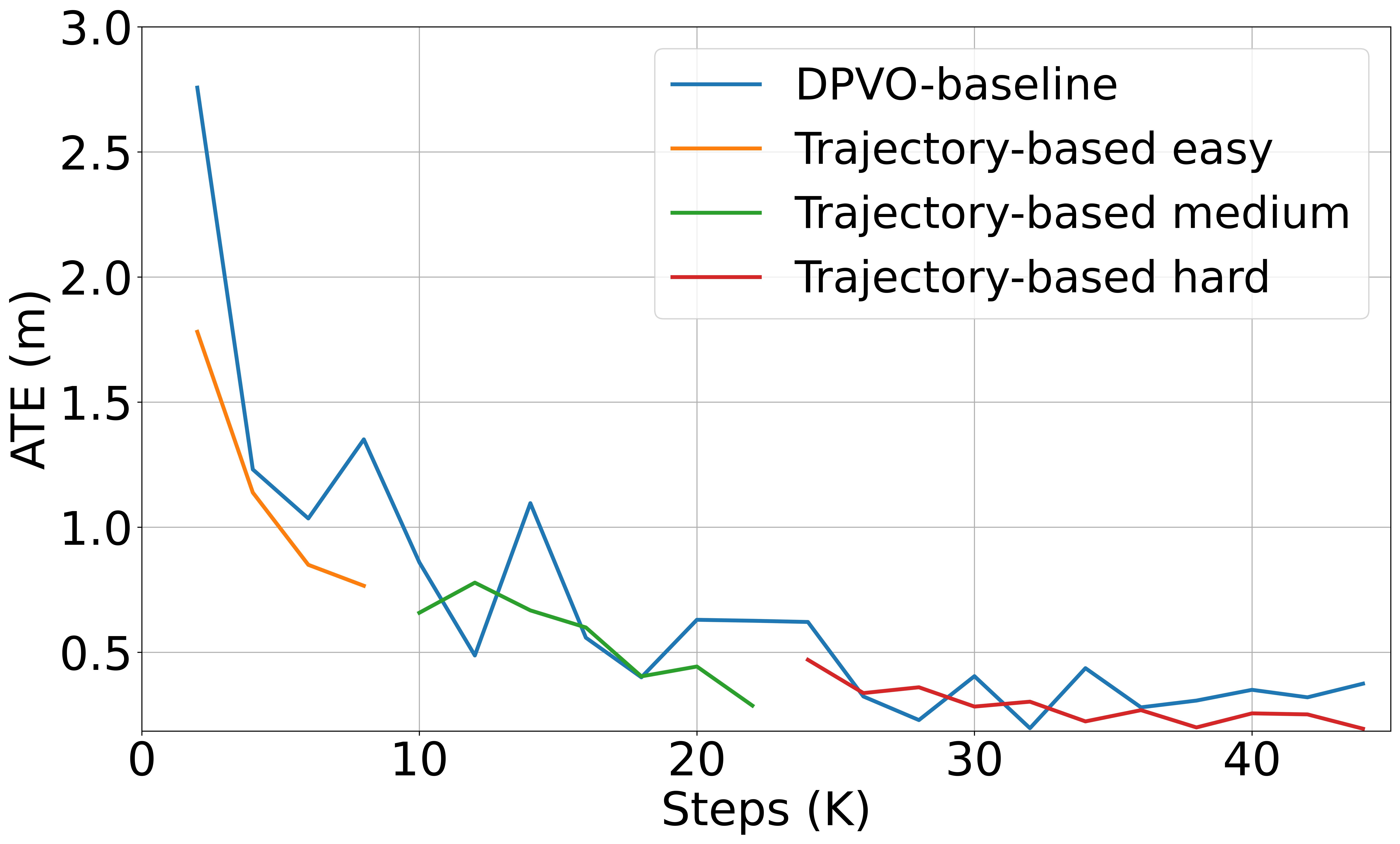}
        \includegraphics[scale=.11]{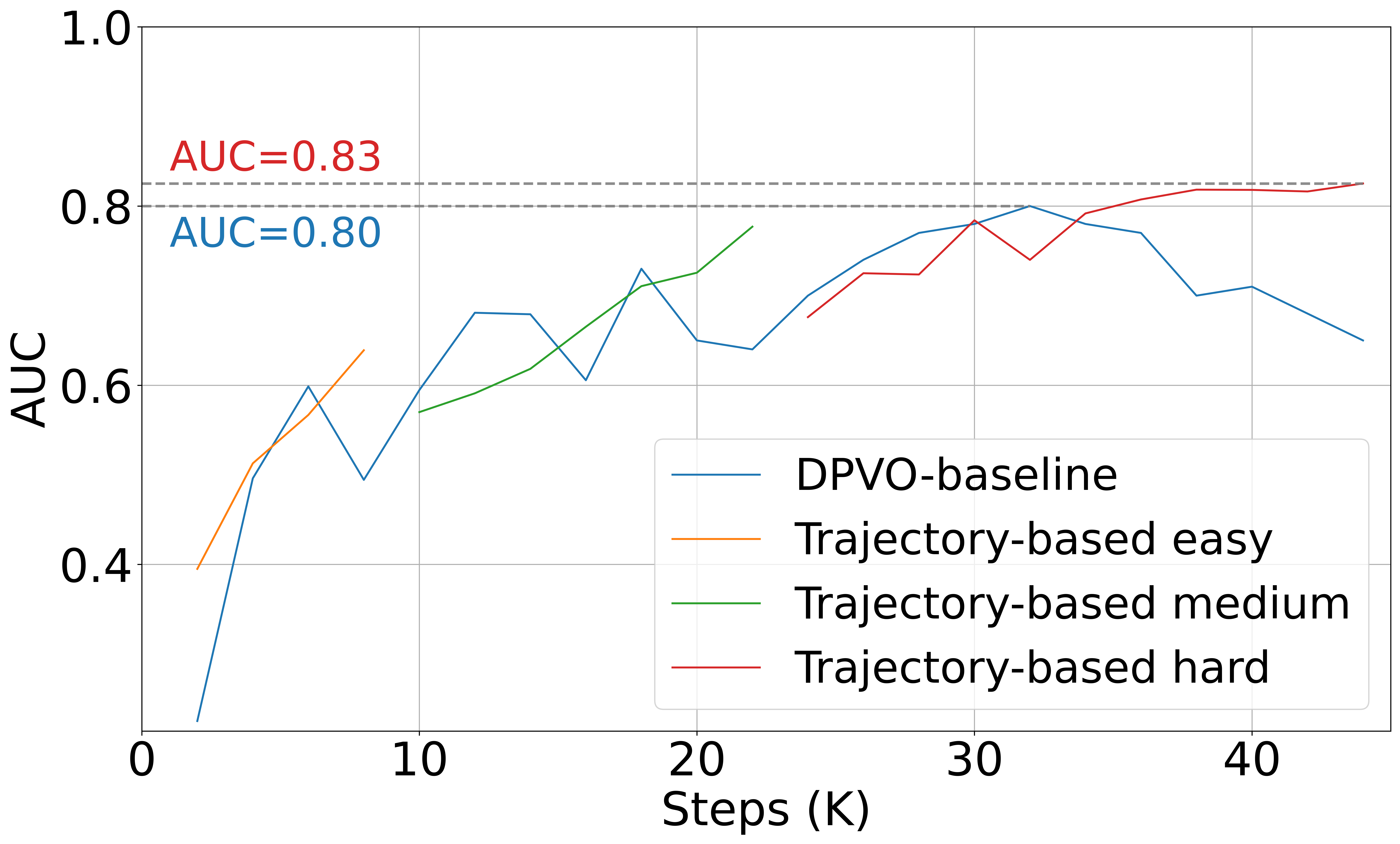}
        \caption{Model validation set performance metrics (ATE and AUC) during training with Trajectory-Based Curriculum Learning strategy. The hard curriculum learning phase (red) improve DPVO baseline results (blue) with ATE=0.17 and AUC=0.83, while achieving comparable validation performance with only the easy and medium difficulty levels samples.}
        \label{fig:trajectoryBasedCLTrain}
	\end{center}
    \end{figure}

	\vspace{0.5em}
	\noindent
	\textbf{Trajectory-Based}: For our Trajectory-Based curriculum learning, we created three distinct trajectory subsets corresponding to different difficulty levels (defined in \mytableref{tab:diffScoresThresholds}). Training proceeded sequentially through these difficulty stages, with each stage initialized from the best checkpoint of the previous stage, halting before overfitting occurred. \autoref{fig:trajectoryBasedCLTrain} shows both AUC and average ATE of the validation set observed during training of the three-phase trajectory-Based strategy (CL-DPVO-Trajectory-Based) compared to the baseline DPVO (blue line). The medium difficulty phase (green) achieves near baseline performance at step 22k (AUC=0.78, ATE=0.22), while the hard difficulty phase continues improving beyond step 32k where baseline overfits, ultimately reaching AUC=0.83 and ATE=0.17. The results demonstrate the effectiveness of curriculum learning, which not only prevents overfitting beyond the baseline's limitations but also achieves superior validation performance.

    Notably, it reaches comparable performance at step 22k, representing a 31{\%} reduction in training time compared to the baseline DPVO. This suggests that the hard samples in the original dataset may not contribute significantly to the model's overall learning capability when used in conventional random training progression. However, through the implementation of curriculum learning with distinct difficulty phases, we were able to effectively incorporate the more challenging trajectories, leading to improved validation performance.
	
\begin{figure}[ht!]
        \begin{center}
    	\includegraphics[scale=.11]{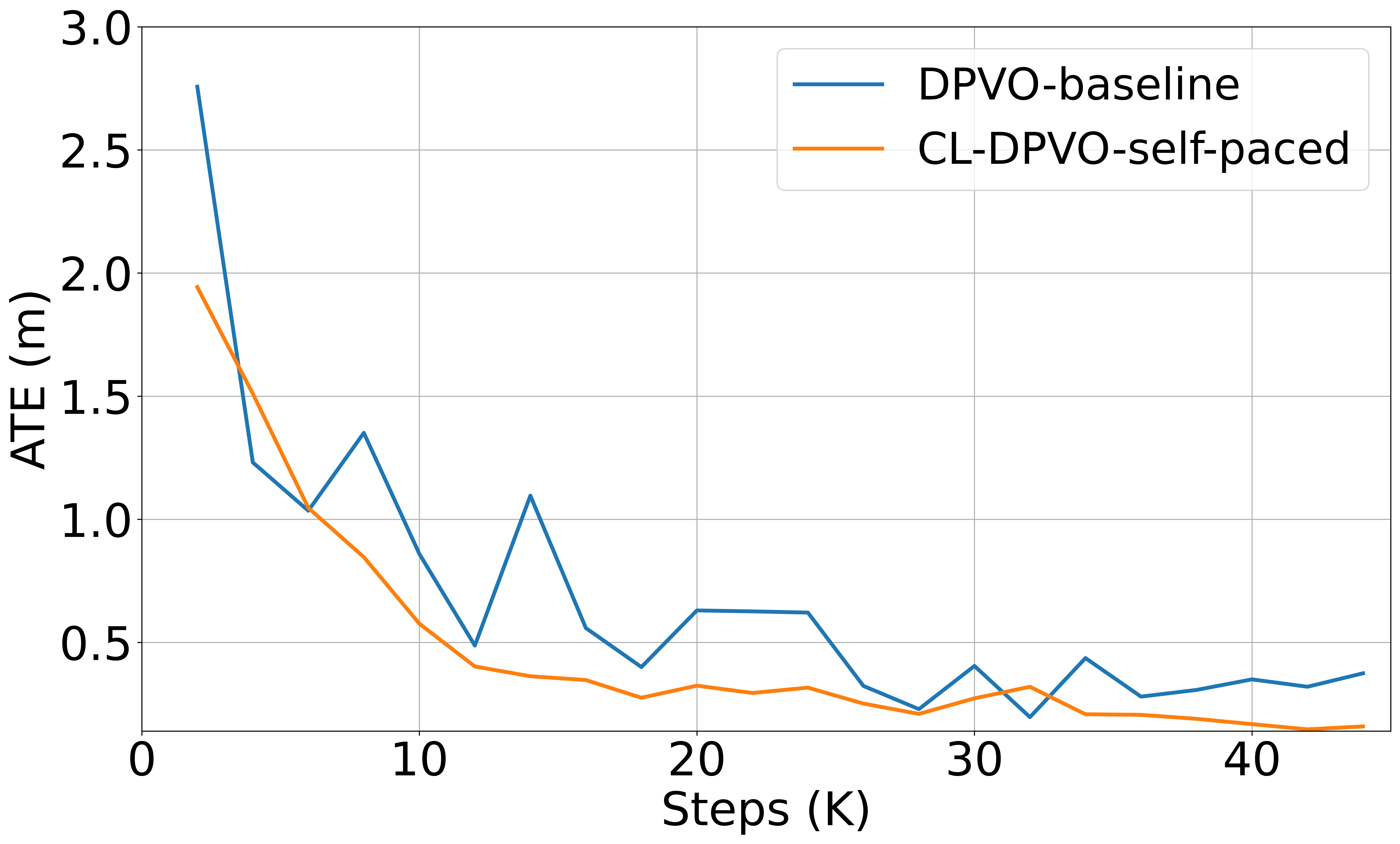}
        \includegraphics[scale=.11]{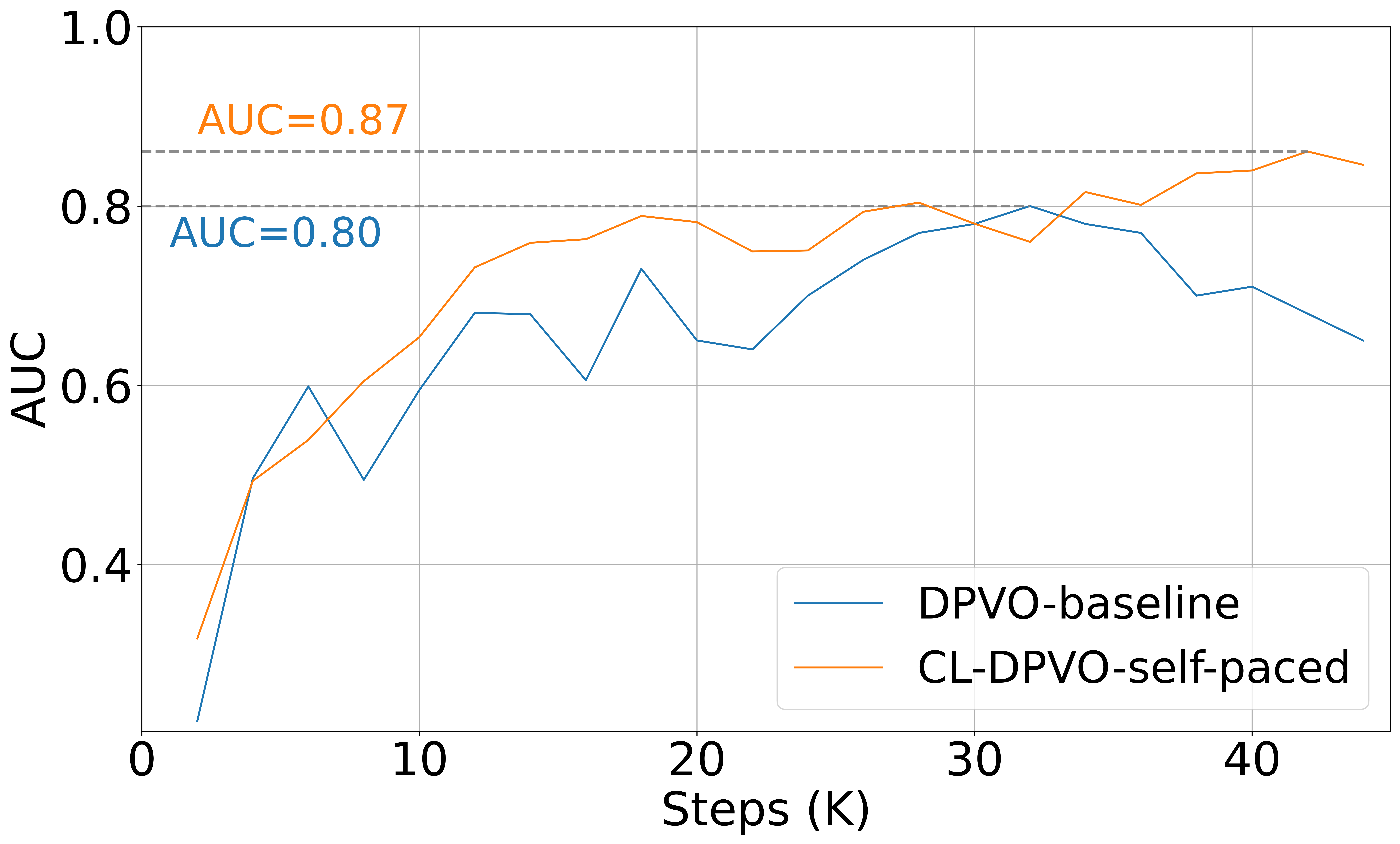}
        \caption{Model validation set performance metrics (ATE and AUC) during training with Self-Paced Curriculum Learning strategy. During early training (step<18k), the Self-Paced approach (orange) exhibits faster and smoother improvements in both ATE and AUC metrics compared to baseline (blue). The method achieves equivalent performance with 47{\%} fewer training steps while reaching the highest AUC (0.87) among all curriculum learning variants.}
        \label{fig:selfPacedCLTrain}
	\end{center}
    \end{figure}

	\vspace{0.5em}
	\noindent
	\textbf{Self-Paced}: For our Self-Paced approach we use a Self-Paced factor (\(\lambda\)=0.1) which demonstrates promising results in \autoref{fig:selfPacedCLTrain}. The validation metrics show earlier convergence than baseline DPVO, achieving comparable performance (AUC=0.8, ATE<0.2) at step 18k, suggesting a 47{\%} reduction in training time. This approach (CL-DPVO-Self-Paced) ultimately reaches an AUC of 0.87 versus the baseline's 0.8. Notably, the early training phase (step<18k) of the Self-Paced method (orange line) demonstrates smoother progression than the baseline approach (blue).

    This improvement stems from the dynamic exponential progression factor in (\ref{eq:selfPacedFactor}) acting as a regularization mechanism, which gradually increases curriculum learning weights when training losses are high and accelerates weight magnitude growth as the model achieves lower loss values, indicating effective training. The adaptive curriculum weights effectively balance the learning process by initially suppressing the impact of difficult samples with high losses, while gradually incorporating them as training progresses. This approach enables effective training to extend beyond step 32k, ultimately achieving state-of-the-art (SOTA) performance of AUC=0.87. 

\begin{figure}[ht!]
        \begin{center}
    	\includegraphics[scale=.11]{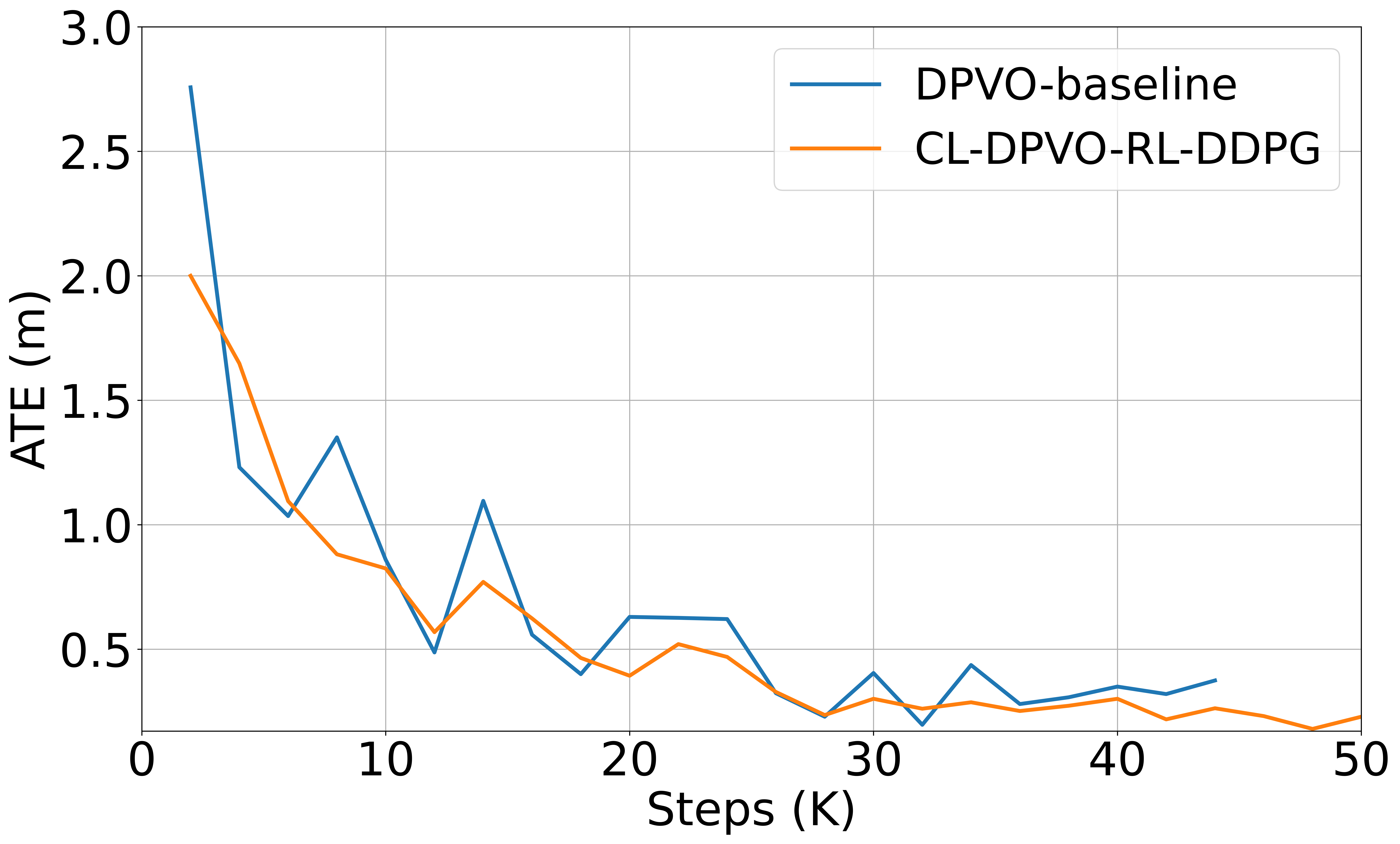}
        \includegraphics[scale=.11]{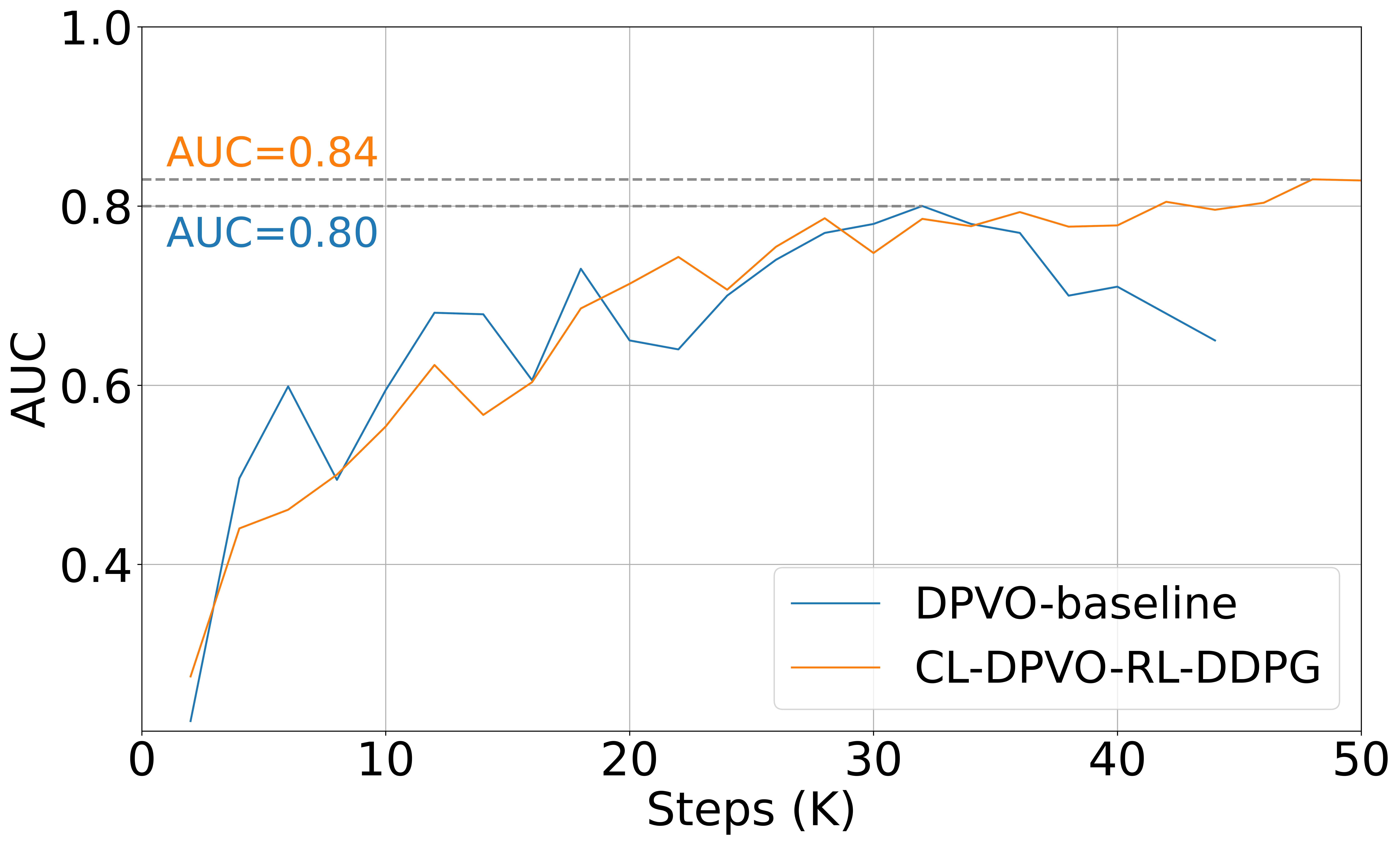}
        \caption{Model validation set performance metrics (ATE and AUC) during training with Reinforcement Learning (DDPG) Curriculum Learning strategy.}
        \label{fig:adaptiveLearningCLTrain}
	\end{center}
    \end{figure}

	\vspace{0.5em}
	\noindent
    \textbf{Adaptive-Learning}: Our RL DDPG agents structured with three-layer actor/critic networks (max width of 64). We maintain separate agent training every k=50 global DPVO training steps for 10 consecutive iterations, using batch size of 64 samples from a 10k-sized replay buffer containing (state, action, reward, next state) tuples. We employ a scaled adaptive noise with scale of 0.1 that reduces near the action space boundaries (0 or 1) to maintain valid actions while balancing exploration and exploitation. The RL strategy (CL-DPVO-RL-DDPG) converges later and slower than previous approaches, reaching AUC=0.84 and ATE=0.15 at step 48k (\autoref{fig:adaptiveLearningCLTrain}). Beyond step 32k, the model shows gradual AUC improvements, eventually exhibiting signs of overfitting after step 48k.

\begin{figure}[ht!]
        \begin{center}
    	\includegraphics[scale=.35]{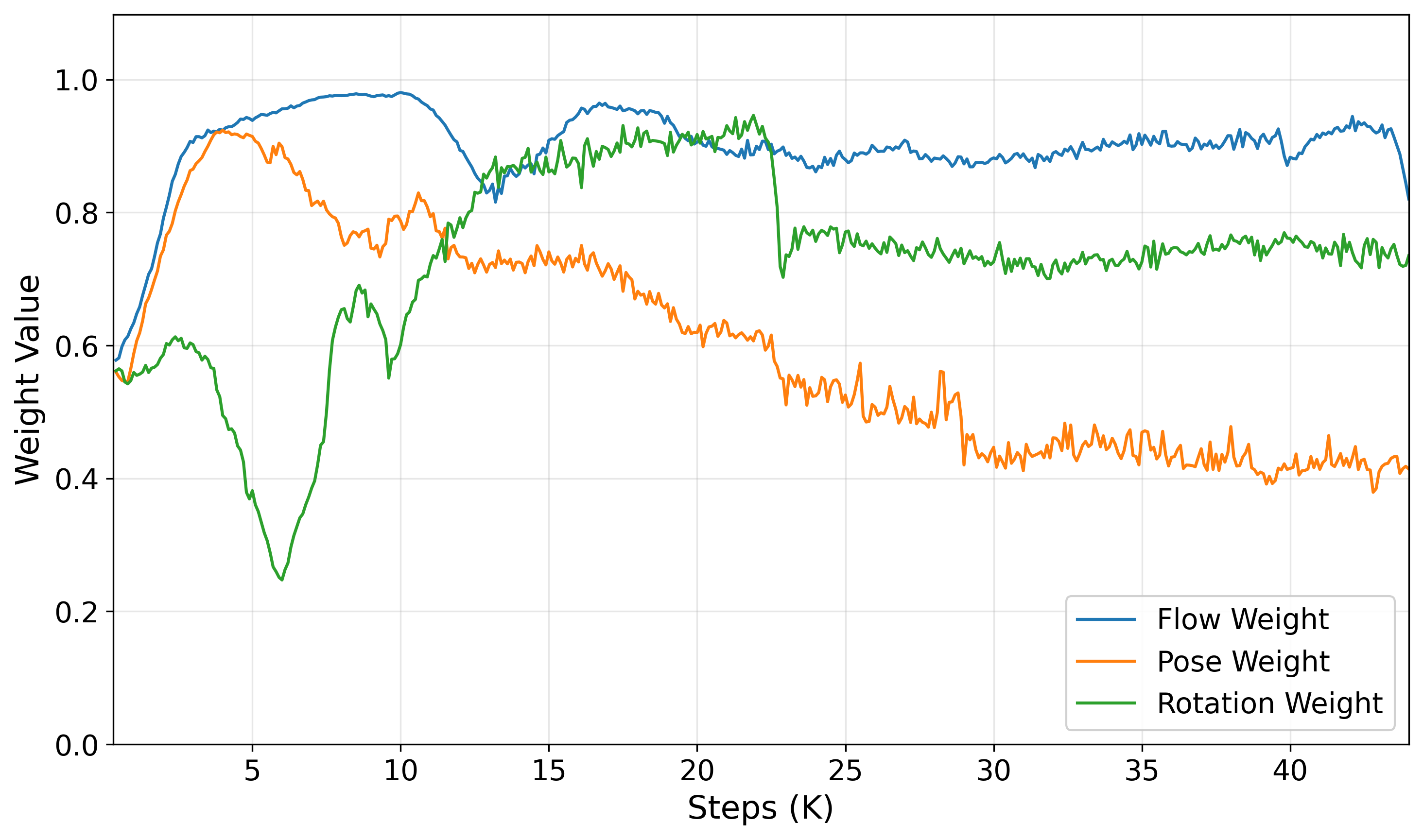}
        \caption{Dynamic weight progression (Flow, Pose, and Rotation) during DPVO training with RL-DDPG Curriculum Learning strategy. Flow weight maintain high values all through the training process alleviates its importance in the overall performance.}
        \label{fig:adaptiveLearningCLTrainWeight}
	\end{center}
    \end{figure}

	\noindent
    This slow convergence likely stems from the DDPG agents' exploration-exploitation balanced nature incorporated into the overall training optimization process. The weight progression shown in \autoref{fig:adaptiveLearningCLTrainWeight} reveals how DDPG agents learn to prioritize different loss components during training. 
	
	Stable weight patterns and clear component prioritization emerge only after step 23k, suggesting a shift from exploration-focused to exploitation-focused behavior. This transition aligns with the observed late convergence in validation metrics. The optical flow weight consistently maintains higher values, suggesting the agents recognize flow estimation's critical role in overall performance. Meanwhile, pose-related weights converge to lower values, though with notable emphasis on rotation components. The learned weight distribution demonstrates the agents' strategy for optimize the balance between flow accuracy and pose estimation, with specific emphasis on the embedded rotational motion elements within pose estimation.

\begin{figure}[ht!]
        \begin{center}
    	\includegraphics[scale=.35]{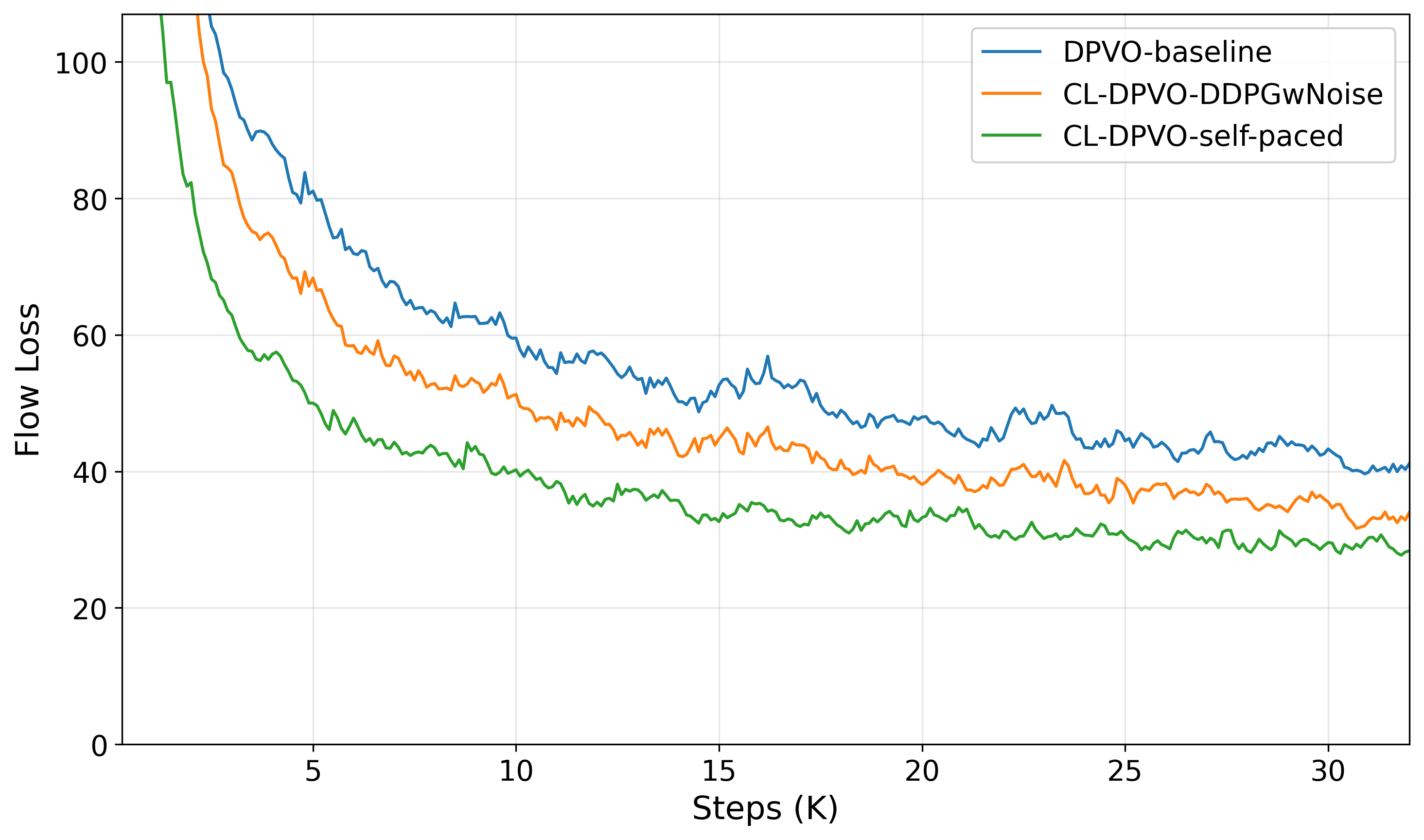}
        \caption{Flow loss progression comparing dynamic curriculum learning strategies (RL-DDPG and Self-Paced) and baseline, with smoothed results for trend visibility. RL-DDPG and Self-Paced significantly outperform the baseline model.}
        \label{fig:adaptiveLearningCLTrainFlow}
        \end{center}
    \end{figure}

    Following the flow loss prominence revealed in \autoref{fig:adaptiveLearningCLTrainWeight}, we examine flow loss training progression across our adaptive CL methods. As shown in \autoref{fig:adaptiveLearningCLTrainFlow}, both adaptive approaches (RL-DDPG and Self-Paced) achieve substantially lower flow loss compared to the baseline, with the Self-Paced method demonstrating superior flow optimization capabilities. This performance correlates with the DDPG agents' learning patterns (\autoref{fig:adaptiveLearningCLTrainWeight}), characterized by consistently elevated flow weights during training.

\begin{table*}[!t]
    \begin{center}
    \caption{Results on the TartanAir monocular test split from the ECCV 2020 SLAM competition. Results are reported as ATE with scale alignment. For our method, we report the median of 5 runs. Top performing (without global optimization/loop closure) method marked in bold with second best underlined. Methods marked with (*) use global optimization / loop closure. Methods marked (\textsuperscript{2}) are image based, where (\textsuperscript{3}) use image-event methods.}
	\renewcommand{\arraystretch}{1.2}
	\label{tab:tartanAirTestSplitAte}
    \begin{tabular}{|c|c|@{\hspace{3pt}}c@{\hspace{3pt}}c@{\hspace{3pt}}c@{\hspace{3pt}}c@{\hspace{3pt}}c@{\hspace{3pt}}c@{\hspace{3pt}}c@{\hspace{3pt}}c@{\hspace{3pt}}c@{\hspace{3pt}}c@{\hspace{3pt}}c@{\hspace{3pt}}c@{\hspace{3pt}}c@{\hspace{3pt}}c@{\hspace{3pt}}c|c|}
		\hline
        \hline
        & ME & ME & ME & ME & ME & ME & ME & ME & MH & MH & MH & MH & MH & MH & MH & MH & \\
        & 000 & 001 & 002 & 003 & 004 & 005 & 006 & 007 & 000 & 001 & 002 & 003 & 004 & 005 & 006 & 007 & Avg \\
		\hline
		ORB-SLAM3* \cite{campos2021orb} & 13.61 & 16.86 & 20.57 & 16.00 & 22.27 & 9.28 & 21.61 & 7.74 & 14.44 & 2.92 & 13.51 & 8.18 & 2.59 & 21.91 & 11.70 & 25.88 & 14.38 \\
		COLMAP* \cite{7780814} & 15.20 & 5.58 & 10.86 & 3.93 & 2.62 & 14.78 & 7.00 & 18.47 & 12.26 & 13.45 & 13.45 & 20.95 & 24.97 & 16.79 & 7.01 & 7.97 & 12.50 \\
		DROID-SLAM* \cite{teed2021droid} & 0.17 & 0.06 & 0.36 & 0.87 & 1.14 & 0.13 & 1.13 & \textbf{0.06} & \textbf{0.08} & 0.05 & \underline{0.04} & \textbf{0.02} & \textbf{0.01} & 0.68 & 0.30 & 0.07 & 0.33 \\
		\hline
		DROID-VO² \cite{teed2021droid} & 0.22 & 0.15 & 0.24 & 1.27 & 1.04 & 0.14 & 1.32 & 0.77 & 0.32 & 0.13 & 0.08 & 0.09 & 1.52 & 0.69 & 0.39 & 0.97 & 0.58 \\
		DPVO² \cite{lipson2024deep} & 0.16 & 0.11 & 0.11 & 0.66 & 0.31 & 0.14 & 0.30 & 0.13 & 0.21 & 0.04 & \underline{0.04} & 0.08 & 0.58 & \textbf{0.17} & 0.11 & 0.15 & 0.21 \\
		RAMP-VO³ \cite{pellerito2024deep} & 0.20 & \textbf{0.04} & \textbf{0.10} & 0.46 & \textbf{0.16} & 0.13 & \textbf{0.12} & 0.12 & 0.36 & 0.06 & \underline{0.04} & 0.04 & 0.41 & 0.25 & 0.11 & \textbf{0.07} & \underline{0.17} \\
		\hline
        CL-DPVO (Trajectory-Based) & 0.13 & \underline{0.05} & \textbf{0.10} & 0.40 & 0.24 & \textbf{0.06} & 0.21 & \underline{0.10} & 0.40 & \textbf{0.02} & \textbf{0.03} & \underline{0.03} & 0.54 & 0.42 & 0.17 & 0.09 & 0.19 \\
		CL-DPVO (RL-DDPG) & \underline{0.12} & \underline{0.05} & \underline{0.11} & \textbf{0.35} & 0.45 & \underline{0.09} & 0.16 & 0.11 & 0.45 & \underline{0.03} & \underline{0.04} & \underline{0.03} & 0.35 & 0.32 & \textbf{0.09} & \underline{0.08} & 0.18 \\
		CL-DPVO (Self-Paced) & \textbf{0.10} & \underline{0.05} & 0.14 & \underline{0.38} & \underline{0.19} & \textbf{0.06} & 0.34 & 0.11 & 0.26 & \underline{0.03} & 0.05 & \textbf{0.02} & \underline{0.18} & \underline{0.21} & \underline{0.10} & 0.10 & \textbf{0.14} \\
        \hline
    \end{tabular}
    \end{center}
\end{table*}

\subsection{Validation}

\begin{figure}[ht!]
        \begin{center}
    	\includegraphics[scale=.35]{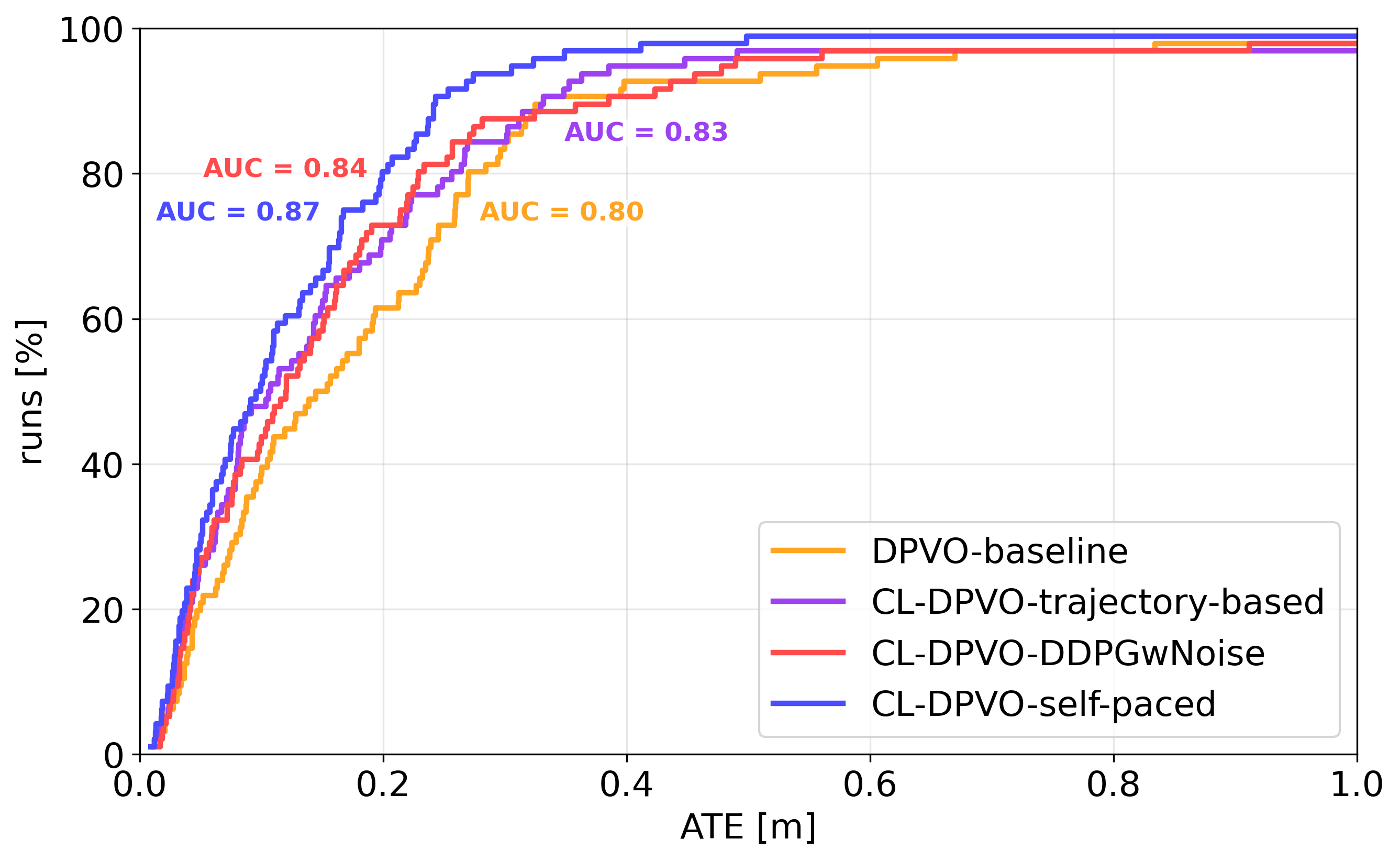}
        \caption{The comparison of the Area Under the Curve (AUC) for the validation set across all DPVO strategies involves averaging the results from three runs for each strategy model.}
        \label{fig:trajectoryBasedCLTrainAuc}
        \end{center}
    \end{figure}

    Following DPVO's evaluation protocol in \cite{lipson2024deep}, we assess our methods on the same 32-sequence validation split, running each sequence three times for consistent comparison. \autoref{fig:trajectoryBasedCLTrainAuc} presents the performance of our strategies within the [0, 1]m error window. The CL-DPVO-Self-Paced demonstrates the strongest performance with an AUC of 0.87, followed by the RL-DDPG based approach at 0.84, and the Trajectory-Based method at 0.83 - all showing improvements over the baseline DPVO's AUC of 0.80.

\subsection{Comparison with State of the Art}

\begin{figure}[ht!]
        \begin{center}
    	\includegraphics[scale=.23]{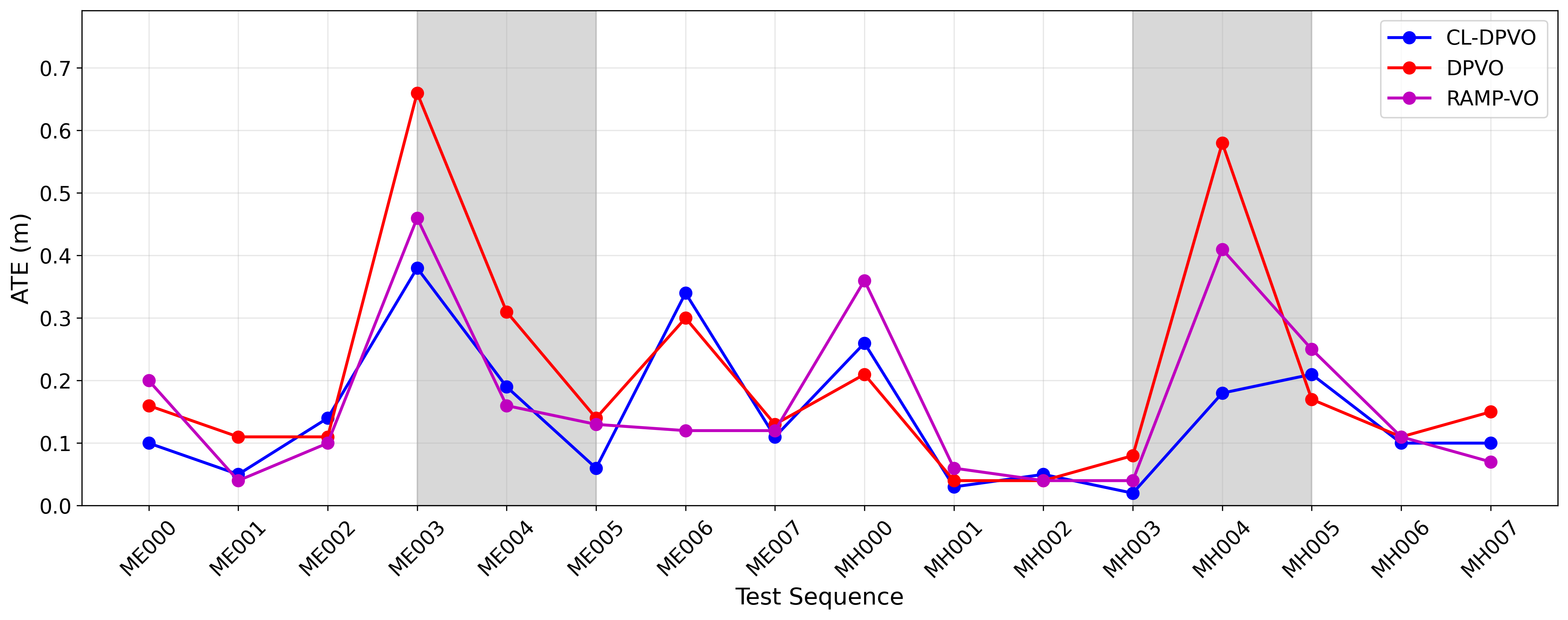}
        \caption{Visualizing TartanAir test split results for CL-DPVO, DPVO, and RAMP-VO. ME (easy motion sequences) and MH (hard motion sequences). Shaded areas are where ATE reduction is most evident.}
        \label{fig:tartanAirTestSplitAte}
        \end{center}
    \end{figure}

	\noindent
    \textbf{TartanAir Test Split}: We compare our CL strategies models with state-of-the-art (SOTA) methods on the TartanAir test-split from the ECCV 2020 SLAM competition, including improved image and event mixture methods. We follow the evaluation in \cite{lipson2024deep}, \cite{pellerito2024deep} and \cite{klenk2024deep} and select ORB-SLAM3 \cite{campos2021orb}, COLMAP \cite{7780814}, \cite{schonberger2016pixelwise}, DROID \cite{teed2021droid} and DPVO \cite{lipson2024deep} as image-only baselines, while we use RAMP-VO \cite{pellerito2024deep} for comparison against the latest updated state-of-the-art (SOTA) image-event method. As in \cite{lipson2024deep} we report the ATE[m] of the median of 5 runs with scale alignment. We use the same default DPVO model configuration as in \cite{lipson2024deep} with 96 patches per frame and 10 frame optimization window. Results for ECCV 2020 competition are available in \mytableref{tab:tartanAirTestSplitAte}. The CL-DPVO (Self-Paced) improves average ATE performance compared to all other image, event and image-event based state-of-the-art methods, outperforming RAMP-VO \cite{pellerito2024deep} by 18{\%} (0.17m to 0.14m) and a 33{\%} relative improvement from the baseline DPVO (0.21m to 0.14m). It shows robust and consistent performance across all scenarios where 13/16 sequences stay below 0.20m and a total narrow range of 0.02-0.38m. The other two CL strategies models, Trajectory-Based and RL-DDPG, improved baseline average ATE by 9{\%} and 14{\%} respectively. In general, the CL-DPVO self-Paced strategy model is able to outperform all other state-of-the-art methods in most cases, including ones using loop closure like DROID-SLAM \cite{teed2021droid}. As shown in \autoref{fig:tartanAirTestSplitAte}, our CL-DPVO achieves its most significant error reductions in sequences ME03-ME05 and MH03-MH05 (highlighted in shaded regions). These sequences, which produce peak ATE values in both DPVO and RAMP-VO, show markedly improved performance under our approach, especially in sequence MH04.

	The robustness of our CL-DPVO method is further demonstrated by analyzing the standard deviation (Std) of the ATE, as presented in \mytableref{tab:TartanAir-Std}. We calculated the Std for sequences ME00-ME07, classified as having easy motion patterns \cite{wang2020tartanair}, and MH00-MH07, classified as having hard motion patterns, as well as the overall global Std in the TartanAir test set.
    \begin{table}[!hbt]

		\begin{center}
		\renewcommand{\arraystretch}{1.2}
		\caption{Motion Pattern Easy (ME), Motion Pattern Hard (MH) and global ATE standard-deviation (Std) of the TartanAir test split sequences.}
		\label{tab:TartanAir-Std}

		\begin{tabular}{|c|c|c|c|}

			\hline
			\hline
			& ME Std[m] & MH Std[m] & Global Std[m] \\
			
            \hline
			DROID-VO \cite{teed2021droid} & 0.483 & 0.477 & 0.483 \\
            RAMP-VO \cite{pellerito2024deep} & 0.119	& 0.141	& 0.1312 \\
            DPVO \cite{lipson2024deep} & 0.176 & 0.164 & 0.173 \\
			\hline
            CL-DPVO (Self-Paced) & \textbf{0.117} & \textbf{0.083} & \textbf{0.105} \\
			\hline
		\end{tabular}
		\end{center}
	\end{table}

	Our top-performing model, CL-DPVO-Self-Paced, is compared against the leading models listed in \mytableref{tab:tartanAirTestSplitAte}, including DROID-VO, RAMP-VO, and DPVO.

    Our CL-DPVO-Self-Paced demonstrates superior consistency by maintaining the lowest global standard deviation in Absolute Trajectory Error (ATE). The improvement is particularly pronounced in the challenging hard sequences (MH), where it reduces standard deviation by 49{\%} compared to the baseline DPVO and 41{\%} compared to RAMP-VO. Globally, our approach achieves a 39{\%} reduction in standard deviation compared to DPVO and a 19{\%} improvement over the current state-of-the-art RAMP-VO, indicating significantly more stable performance across all sequences.
	
	\vspace{0.5em}
	\noindent
    \textbf{EuRoC MAV \cite{burri2016euroc}}: We use our 3 CL strategies models trained on TartanAir and benchmark on the EuRoC MAV dataset. \mytableref{tab:EuRoCTestSplitAte} displays the sequence-specific and average ATE for the test split, benchmarking our CL-DPVO against other visual odometry techniques, including SVO \cite{forster2014svo}, DSO \cite{engel2017direct}, and DROID-VO (derived from DROID-SLAM \cite{teed2021droid} without global optimization techniques). Results are taken from \cite{lipson2024deep}. Among the evaluated strategies, the Self-Paced CL-DPVO emerges as the top performer, achieving a 13{\%} reduction in average ATE compared to the baseline DPVO, and surpassing the next best method, DROID-VO, by 51{\%}. Both best performing methods, CL-DPVO-Self-Paced (in bold) and CL-DPVO-RL-DDPG (in underline), are able to outperform the other state-of-the-art methods in most cases.

	\begin{table}[!hbt]
		
		\begin{center}
			
			\caption{Results of the Avg. ATE[m] on the EuRoC test split monocular SLAM dataset}
			\label{tab:EuRoCTestSplitAte}
			\renewcommand{\arraystretch}{1.3}
			\begin{tabular}{|@{\hspace{3pt}}c@{\hspace{3pt}}|@{\hspace{3pt}}c@{\hspace{3pt}}@{\hspace{3pt}}c@{\hspace{3pt}}@{\hspace{3pt}}c@{\hspace{3pt}}@{\hspace{3pt}}c@{\hspace{3pt}}@{\hspace{3pt}}c@{\hspace{3pt}}|@{\hspace{3pt}}c@{\hspace{3pt}}@{\hspace{3pt}}c@{\hspace{3pt}}@{\hspace{3pt}}c@{\hspace{3pt}}|}
				\hline
				\hline
				& \rotatebox{90}{TartanVO \cite{wang2021tartanvo}} & \rotatebox{90}{SVO \cite{forster2014svo}} & \rotatebox{90}{DSO \cite{engel2017direct}} & \rotatebox{90}{DROID-VO \cite{teed2021droid}}& \rotatebox{90}{DPVO \cite{lipson2024deep}}& \rotatebox{90}{CL-DPVO (Trajectory-Based) }& \rotatebox{90}{CL-DPVO (RL-DDPG)}& \rotatebox{90}{CL-DPVO (Self-Paced)}\\
				\hline
				MH01 & 0.639 & 0.100 & \textbf{0.046} & 0.163 & 0.087 & 0.083 & \underline{0.069} & 0.081 \\
				MH02 & 0.325 & 0.120 & 0.046 & 0.121 & 0.055 & 0.060 & \underline{0.044} & \textbf{0.030} \\
				MH03 & 0.550 & 0.410 & 0.172 & 0.242 & 0.158 & 0.148 & \textbf{0.120} & \underline{0.122} \\
				MH04 & 1.153 & 0.430 & 3.810 & 0.399 & \underline{0.137} & 0.151 & 0.144 & \textbf{0.133} \\
				MH05 & 1.021 & 0.300 & \textbf{0.110} & 0.270 & \underline{0.114} & 0.123 & 0.115 & \underline{0.114} \\
				V101 & 0.447 & 0.070 & 0.089 & 0.103 & \textbf{0.050} & \underline{0.051} & 0.053 & \underline{0.051} \\
				V102 & 0.389 & 0.210 & \textbf{0.107} & 0.165 & 0.140 & 0.146 & 0.145 & \underline{0.118} \\
				V103 & 0.622 & -	 & 0.903 & 0.158 & 0.086 & \textbf{0.055} & \underline{0.061} & 0.063 \\
				V201 & 0.433 & 0.110 & \textbf{0.044} & 0.102 & \underline{0.057} & 0.060 & 0.078 & 0.065 \\
				V202 & 0.749 & 0.110 & 0.132 & 0.115 & \underline{0.049} & 0.056 & 0.052 & \textbf{0.045} \\
				V203 & 1.152 & 1.080 & 1.152 & 0.204 & 0.211 & 0.219 & \textbf{0.149} & \underline{0.178} \\
				\hline
				Avg & 0.680 & 0.294 & 0.601 & 0.186 & 0.105 & 0.105 & \underline{0.094} & \textbf{0.091} \\
				\hline
			\end{tabular}
		\end{center}
	\end{table}

	\vspace{0.5em}
	\noindent
	\textbf{TUM-RGBD \cite{sturm2012benchmark}}: In \mytableref{tab:TUM-RGBDTestSplitAte}, we benchmark our CL strategies on the TUM-RGBD dataset, comparing them against DROID-VO \cite{teed2021droid} and DPVO \cite{lipson2024deep}. Our evaluation focuses solely on visual-only monocular methods, consistent with the approach in \cite{teed2021droid} for this dataset. This benchmark tests motion tracking in an indoor setting with erratic camera movements and substantial motion blur. According to \cite{lipson2024deep}, traditional methods like ORB-SLAM \cite{campos2021orb} and DSO \cite{engel2017direct} perform adequately on specific sequences but are prone to frequent catastrophic failures. We focus on methods capable of providing results for all sequences in the test split. We follow the evaluation settings provided by DROID-SLAM \cite{teed2021droid} and calculate the median of 5 runs. Like the baseline DPVO, our CL models demonstrate robustness to sequence failures, with the Self-Paced model showing a 9{\%} reduction in avarage ATE compared to the baseline DPVO.
    
	\begin{table}[!hbt]
    	
    	\begin{center}
    		
    		\caption{Results (ATE) on the freiburg1 set of TUM-RGBD. We use monocular visual odometry only (No method uses stereo or sensor depth) and identical evaluation setting as in DROID-SLAM.}
    		\label{tab:TUM-RGBDTestSplitAte}
    		\renewcommand{\arraystretch}{1.2}
			\begin{tabular}{|@{\hspace{3pt}}c@{\hspace{3pt}}|@{\hspace{3pt}}c@{\hspace{3pt}}@{\hspace{3pt}}c@{\hspace{3pt}}|@{\hspace{3pt}}c@{\hspace{3pt}}@{\hspace{3pt}}c@{\hspace{3pt}}@{\hspace{3pt}}c@{\hspace{3pt}}|}
    			\hline
    			\hline
    			& \rotatebox{90}{DROID-VO \cite{teed2021droid}}& \rotatebox{90}{DPVO \cite{lipson2024deep}}& \rotatebox{90}{CL-DPVO (Trajectory-Based) }& \rotatebox{90}{CL-DPVO (RL-DDPG)}& \rotatebox{90}{CL-DPVO (Self-Paced)}\\
    			\hline
    			360 & 0.161 & 0.135 & 0.130 & \underline{0.127} & \textbf{0.122} \\
    			desk & \underline{0.028} & 0.038 & 0.050 & 0.061 & \textbf{0.025} \\
    			desk2 & 0.099 & \underline{0.048} & \textbf{0.041} & 0.049 & \underline{0.048} \\
    			floor & \textbf{0.033} & 0.040 & 0.049 & 0.046 & \underline{0.036} \\
    			plant & 0.028 & 0.036 & \underline{0.026} & \textbf{0.023} & 0.027 \\
    			room & \textbf{0.327} & 0.394 & 0.393 & \underline{0.329} & 0.351 \\
    			rpy & \textbf{0.028} & 0.034 & 0.045 & \underline{0.031} & \underline{0.031} \\
    			teddy & 0.169 & 0.064 & 0.074 & \textbf{0.048} & \underline{0.056} \\
    			xyz & 0.013 & \underline{0.012} & \textbf{0.011} & \underline{0.012} & 0.013 \\
    			\hline
    			Avg & 0.098 & 0.089 & 0.091 & \underline{0.081} & \textbf{0.079} \\
    			\hline
    		\end{tabular}
    	\end{center}
    \end{table}
	\noindent
    Both Self-Paced and RL-DDPG strategies outperform the baseline DPVO, and DROID-VO, with the Self-Paced model achieving the best performance.

    \vspace{0.5em}
	\noindent
    \textbf{ICL-NUIM \cite{handa2014benchmark}}: In \mytableref{tab:ICL-NUIMTestSplitAte}, we assess our CL-DPVO models using the ICL-NUIM SLAM benchmark, contrasting them with leading visual odometry and SLAM techniques such as SVO \cite{forster2014svo}, DSO \cite{engel2017direct}, DROID-SLAM \cite{teed2021droid}, and the baseline DPVO. We follow our previous guideline to present only VO methods that succeed in all sequences. The ICL-NUIM dataset is synthetic and designed for evaluating SLAM in indoor settings, characterized by repetitive or monochrome textures like plain white walls and enhanced noise models. All three CL-DPVO variants surpass previous state-of-the-art results. Notably, the Trajectory-Based variant emerges as the leading approach, outperforming both other CL strategies and reducing ATE by 32{\%} compared to the fast variant of DPVO in 6 out of 8 sequences. This is a notable departure from our findings on TartanAir, EuRoC, and TUM-RGBD benchmarks, where the Self-Paced model consistently led performance metrics. The superior performance of the Trajectory-Based approach on ICL-NUIM demonstrates how different curriculum learning strategies can be particularly well-suited for specific challenges - in this case, the explicit Trajectory-Based difficulty progression appears to better handle the precise camera motion requirements needed for accurate surface reconstruction. This finding underscores the versatility of our curriculum learning framework, where different strategies can naturally adapt to and excel in different scenarios, suggesting that the choice of curriculum strategy should be influenced by the specific requirements and characteristics of the target application.
    
    \begin{table}[!hbt]
    	
    	\begin{center}
    		
    		\caption{Results (ATE) on ICL-NUIM SLAM benchmark. Methods marked with (*) use global optimization / loop closure.}
    		\label{tab:ICL-NUIMTestSplitAte}
    		\renewcommand{\arraystretch}{1.3} % Increased spacing between rows
    		\begin{tabular}{|@{\hspace{2pt}}c@{\hspace{2pt}}|@{\hspace{2pt}}c@{\hspace{2pt}}@{\hspace{2pt}}c@{\hspace{2pt}}@{\hspace{2pt}}c@{\hspace{2pt}}@{\hspace{2pt}}c@{\hspace{2pt}}@{\hspace{2pt}}c@{\hspace{2pt}}@{\hspace{2pt}}c@{\hspace{2pt}}@{\hspace{2pt}}c@{\hspace{2pt}}|@{\hspace{2pt}}c@{\hspace{2pt}}@{\hspace{2pt}}c@{\hspace{2pt}}@{\hspace{2pt}}c@{\hspace{2pt}}|}
    			\hline
    			\hline
    			& \rotatebox{90}{DROID-SLAM* \cite{teed2021droid}} & \rotatebox{90}{DROID-VO \cite{teed2021droid}} & \rotatebox{90}{SVO \cite{forster2014svo}} & \rotatebox{90}{DSO \cite{engel2017direct}} & \rotatebox{90}{DSO-Realtime \cite{engel2017direct}}& \rotatebox{90}{DPVO \cite{lipson2024deep}}& \rotatebox{90}{DPVO-Fast \cite{lipson2024deep}} & \rotatebox{90}{CL-DPVO (Trajectory-Based) }& \rotatebox{90}{CL-DPVO (RL-DDPG)}& \rotatebox{90}{CL-DPVO (Self-Paced)}\\
    			\hline
    			lr-kt0 & 0.008 & 0.010 & 0.02 & 0.01 & 0.02 & \underline{0.006} & 0.008 & \underline{0.006} & \textbf{0.005} & 0.007 \\
    			lr-kt1 & 0.027 & 0.123 & 0.07 & 0.02 & 0.03 & 0.006 & 0.007 & \textbf{0.004} & 0.008 & \underline{0.005} \\
    			lr-kt2 & 0.039 & 0.072 & 0.09 & 0.06 & 0.33 & 0.023 & 0.021 & \textbf{0.018} & \underline{0.020} & 0.022 \\
    			lr-kt3 & 0.012 & 0.032 & 0.07 & 0.03 & 0.06 & 0.010 & 0.010 & \textbf{0.005} & \underline{0.006} & \underline{0.006} \\
    			of-kt0 & \underline{0.065} & 0.095 & 0.34 & 0.21 & 0.29 & 0.067 & 0.071 & \textbf{0.007} & \textbf{0.007} & \textbf{0.007} \\
    			of-kt1 & 0.025 & 0.041 & 0.28 & 0.83 & 0.64 & 0.012 & 0.015 & \textbf{0.008} & \textbf{0.008} & \underline{0.009} \\
    			of-kt2 & 0.858 & 0.842 & 0.14 & 0.36 & 0.23 & \underline{0.017} & 0.018 & \textbf{0.015} & 0.026 & 0.024 \\
    			of-kt3 & 0.481 & 0.504 & \textbf{0.08} & 0.64 & 0.46 & 0.635 & 0.593 & \underline{0.442} & 0.459 & 0.466 \\
    			\hline
    			Avg & 0.189 & 0.215 & 0.136 & 0.270 & 0.258 & 0.097 & 0.093 & \textbf{0.063} & \underline{0.067} & 0.068 \\
    			\hline
    		\end{tabular}
    	\end{center}
    \end{table}
    
 \section{Conclusion}
 
 	In this work, we demonstrate the effectiveness of curriculum learning strategies in improving visual odometry performance and robustness. All three approaches - Trajectory-Based, Self-Paced and RL-DDPG - show notable improvements over the baseline DPVO, with the Self-Paced method achieving state-of-the-art performance and outperforming all prior work on the TartanAir ECCV 2020 SLAM competition, EuRoC MAV, and TUM-RGBD SLAM benchmarks. On ICL-NUIM SLAM benchmark our Trajectory-Based CL model outperforms state-of-the-art monocular VO methods including those using optimization techniques. Our Self-Paced method matches the DPVO baseline performance while reducing training time by 47{\%} before reaching superior results further down the training process. Using adaptive off-policy reinforcement learning technique, we uncover the natural equilibrium between visual odometry's core learned components. Our analysis highlights flow estimation as a crucial factor for performance gains and model robustness, while dynamic weight adaptation effectively balances various learning aspects to improve overall results. 
	
	Notably, our comprehensive evaluation across different benchmarks reveals that specific curriculum learning strategies can be particularly well-suited for certain datasets and their unique challenges, as demonstrated by the superior performance of our Trajectory-Based approach on ICL-NUIM's surface reconstruction-focused sequences, suggesting that the choice of curriculum strategy should be influenced by the target application's specific requirements. Although demonstrated with DPVO, our curriculum learning framework represents a general methodology that can be integrated into various visual odometry architectures to enhance their real-world performance and robustness.
 	
\nocite{*}
\printbibliography[
heading=bibintoc,
title={References}
]

@inproceedings{10.1145/1553374.1553380,
	author = {Bengio, Yoshua and Louradour, J\'{e}r\^{o}me and Collobert, Ronan and Weston, Jason},
	title = {Curriculum learning},
	year = {2009},
	publisher = {Association for Computing Machinery},
	address = {New York, NY, USA},
	booktitle = {Proceedings of the 26th Annual International Conference on Machine Learning},
	pages = {41–48},
	numpages = {8},
	location = {Montreal, Quebec, Canada},
	series = {ICML '09}
}

@inproceedings{weinshall2018curriculum,
  title={Curriculum learning by transfer learning: Theory and experiments with deep networks},
  author={Weinshall, Daphna and Cohen, Gad and Amir, Dan},
  booktitle={International conference on machine learning},
  pages={5238--5246},
  year={2018},
  organization={PMLR}
}

@inproceedings{lipson2024deep,
  title={Deep patch visual odometry},
  author={Teed, Zachary and Lipson, Lahav and Deng, Jia},
  journal={Advances in Neural Information Processing Systems},
  volume={36},
  year={2024}
}

@inproceedings{pellerito2024deep,
  title={Deep Visual Odometry with Events and Frames},
  author={Pellerito, Roberto and Cannici, Marco and Gehrig, Daniel and Belhadj, Joris and Dubois-Matra, Olivier and Casasco, Massimo and Scaramuzza, Davide},
  booktitle={Proceedings of the IEEE/RSJ International Conference on Intelligent Robots and Systems (IROS. IEEE)},
  year={2024}
}

@inproceedings{saputra2019learning,
  title={Learning monocular visual odometry through geometry-aware curriculum learning},
  author={Saputra, Muhamad Risqi U and De Gusmao, Pedro PB and Wang, Sen and Markham, Andrew and Trigoni, Niki},
  booktitle={2019 international conference on robotics and automation (ICRA)},
  pages={3549--3555},
  year={2019},
  organization={IEEE}
}

@inproceedings{hacohen2019power,
  title={On the power of curriculum learning in training deep networks},
  author={Hacohen, Guy and Weinshall, Daphna},
  booktitle={International conference on machine learning},
  pages={2535--2544},
  year={2019},
  organization={PMLR}
}

@article{campos2021orb,
  title={Orb-slam3: An accurate open-source library for visual, visual--inertial, and multimap slam},
  author={Campos, Carlos and Elvira, Richard and Rodr{\'\i}guez, Juan J G{\'o}mez and Montiel, Jos{\'e} MM and Tard{\'o}s, Juan D},
  journal={IEEE Transactions on Robotics},
  volume={37},
  number={6},
  pages={1874--1890},
  year={2021},
  publisher={IEEE}
}

@inproceedings{wang2017deepvo,
  title={Deepvo: Towards end-to-end visual odometry with deep recurrent convolutional neural networks},
  author={Wang, Sen and Clark, Ronald and Wen, Hongkai and Trigoni, Niki},
  booktitle={2017 IEEE international conference on robotics and automation (ICRA)},
  pages={2043--2050},
  year={2017},
  organization={IEEE}
}

@INPROCEEDINGS{1315094,
  author={Nister, D. and Naroditsky, O. and Bergen, J.},
  booktitle={Proceedings of the 2004 IEEE Computer Society Conference on Computer Vision and Pattern Recognition, 2004. CVPR 2004.}, 
  title={Visual odometry}, 
  year={2004},
  volume={1},
  number={},
  pages={I-I},
  doi={10.1109/CVPR.2004.1315094}
}

@article{maimone2007two,
  title={Two years of visual odometry on the mars exploration rovers},
  author={Maimone, Mark and Cheng, Yang and Matthies, Larry},
  journal={Journal of Field Robotics},
  volume={24},
  number={3},
  pages={169--186},
  year={2007},
  publisher={Wiley Online Library}
}

@inproceedings{badino2013visual,
  title={Visual odometry by multi-frame feature integration},
  author={Badino, Hern{\'a}n and Yamamoto, Akihiro and Kanade, Takeo},
  booktitle={Proceedings of the IEEE International Conference on Computer Vision Workshops},
  pages={222--229},
  year={2013}
}

@inproceedings{jiang2015self,
  title={Self-paced curriculum learning},
  author={Jiang, Lu and Meng, Deyu and Zhao, Qian and Shan, Shiguang and Hauptmann, Alexander},
  booktitle={Proceedings of the AAAI Conference on Artificial Intelligence},
  volume={29},
  number={1},
  year={2015}
}

@article{jiang2014self,
  title={Self-paced learning with diversity},
  author={Jiang, Lu and Meng, Deyu and Yu, Shoou-I and Lan, Zhenzhong and Shan, Shiguang and Hauptmann, Alexander},
  journal={Advances in neural information processing systems},
  volume={27},
  year={2014}
}

@inproceedings{kendall2015posenet,
  title={Posenet: A convolutional network for real-time 6-dof camera relocalization},
  author={Kendall, Alex and Grimes, Matthew and Cipolla, Roberto},
  booktitle={Proceedings of the IEEE international conference on computer vision},
  pages={2938--2946},
  year={2015}
}

@inproceedings{klenk2024deep,
  title={Deep event visual odometry},
  author={Klenk, Simon and Motzet, Marvin and Koestler, Lukas and Cremers, Daniel},
  booktitle={2024 International Conference on 3D Vision (3DV)},
  pages={739--749},
  year={2024},
  organization={IEEE}
}

@inproceedings{wang2021tartanvo,
  title={Tartanvo: A generalizable learning-based vo},
  author={Wang, Wenshan and Hu, Yaoyu and Scherer, Sebastian},
  booktitle={Conference on Robot Learning},
  pages={1761--1772},
  year={2021},
  organization={PMLR}
}

@article{wei2016stc,
  title={Stc: A simple to complex framework for weakly-supervised semantic segmentation},
  author={Wei, Yunchao and Liang, Xiaodan and Chen, Yunpeng and Shen, Xiaohui and Cheng, Ming-Ming and Feng, Jiashi and Zhao, Yao and Yan, Shuicheng},
  journal={IEEE transactions on pattern analysis and machine intelligence},
  volume={39},
  number={11},
  pages={2314--2320},
  year={2016},
  publisher={IEEE}
}

@inproceedings{8461251,
  author={Li, Ruihao and Wang, Sen and Long, Zhiqiang and Gu, Dongbing},
  booktitle={2018 IEEE International Conference on Robotics and Automation (ICRA)}, 
  title={UnDeepVO: Monocular Visual Odometry Through Unsupervised Deep Learning}, 
  year={2018},
  volume={},
  number={},
  pages={7286-7291},
  doi={10.1109/ICRA.2018.8461251}
}

@article{schuler2015learning,
  title={Learning to deblur},
  author={Schuler, Christian J and Hirsch, Michael and Harmeling, Stefan and Sch{\"o}lkopf, Bernhard},
  journal={IEEE transactions on pattern analysis and machine intelligence},
  volume={38},
  number={7},
  pages={1439--1451},
  year={2015},
  publisher={IEEE}
}

@inproceedings{9025375,
  author={Shermeyer, Jacob and Van Etten, Adam},
  booktitle={2019 IEEE/CVF Conference on Computer Vision and Pattern Recognition Workshops (CVPRW)}, 
  title={The Effects of Super-Resolution on Object Detection Performance in Satellite Imagery}, 
  year={2019},
  volume={},
  number={},
  pages={1432-1441},
  doi={10.1109/CVPRW.2019.00184}
}

@inproceedings{yang2018quality,
  title={Quality classified image analysis with application to face detection and recognition},
  author={Yang, Fei and Zhang, Qian and Wang, Miaohui and Qiu, Guoping},
  booktitle={2018 24th International Conference on Pattern Recognition (ICPR)},
  pages={2863--2868},
  year={2018},
  organization={IEEE}
}

@article{dodge2018quality,
  title={Quality robust mixtures of deep neural networks},
  author={Dodge, Samuel F and Karam, Lina J},
  journal={IEEE Transactions on Image Processing},
  volume={27},
  number={11},
  pages={5553--5562},
  year={2018},
  publisher={IEEE}
}

@inproceedings{dodge2016understanding,
  title={Understanding how image quality affects deep neural networks},
  author={Dodge, Samuel and Karam, Lina},
  booktitle={2016 eighth international conference on quality of multimedia experience (QoMEX)},
  pages={1--6},
  year={2016},
  organization={IEEE}
}

@inproceedings{liu2017quality,
  title={Quality aware network for set to set recognition},
  author={Liu, Yu and Yan, Junjie and Ouyang, Wanli},
  booktitle={Proceedings of the IEEE conference on computer vision and pattern recognition},
  pages={5790--5799},
  year={2017}
}

@inproceedings{wang2020tartanair,
  title={TartanAir: A dataset to push the limits of visual slam},
  author={Wang, Wenshan and Zhu, Delong and Wang, Xiangwei and Hu, Yaoyu and Qiu, Yuheng and Wang, Chen and Hu, Yafei and Kapoor, Ashish and Scherer, Sebastian},
  booktitle={2020 IEEE/RSJ International Conference on Intelligent Robots and Systems (IROS)},
  pages={4909--4916},
  year={2020},
  organization={IEEE}
}

@article{teed2021droid,
  title={Droid-slam: Deep visual slam for monocular, stereo, and rgb-d cameras},
  author={Teed, Zachary and Deng, Jia},
  journal={Advances in neural information processing systems},
  volume={34},
  pages={16558--16569},
  year={2021}
}

@article{qin2019general,
  title={A general optimization-based framework for global pose estimation with multiple sensors},
  author={Qin, Tong and Cao, Shaozu and Pan, Jie and Shen, Shaojie},
  journal={arXiv preprint arXiv:1901.03642},
  year={2019}
}

@inproceedings{clark2017vinet,
  title={Vinet: Visual-inertial odometry as a sequence-to-sequence learning problem},
  author={Clark, Ronald and Wang, Sen and Wen, Hongkai and Markham, Andrew and Trigoni, Niki},
  booktitle={Proceedings of the AAAI conference on artificial intelligence},
  volume={31},
  number={1},
  year={2017}
}

@article{umeyama1991least,
  title={Least-squares estimation of transformation parameters between two point patterns},
  author={Umeyama, Shinji},
  journal={IEEE Transactions on Pattern Analysis \& Machine Intelligence},
  volume={13},
  number={04},
  pages={376--380},
  year={1991},
  publisher={IEEE Computer Society}
}

@inproceedings{7780814,
  author={Schönberger, Johannes L. and Frahm, Jan-Michael},
  booktitle={2016 IEEE Conference on Computer Vision and Pattern Recognition (CVPR)}, 
  title={Structure-from-Motion Revisited}, 
  year={2016},
  volume={},
  number={},
  pages={4104-4113},
  doi={10.1109/CVPR.2016.445}
}

@inproceedings{schonberger2016pixelwise,
  title={Pixelwise view selection for unstructured multi-view stereo},
  author={Sch{\"o}nberger, Johannes L and Zheng, Enliang and Frahm, Jan-Michael and Pollefeys, Marc},
  booktitle={Computer Vision--ECCV 2016: 14th European Conference, Amsterdam, The Netherlands, October 11-14, 2016, Proceedings, Part III 14},
  pages={501--518},
  year={2016},
  organization={Springer}
}

@inproceedings{hidalgo2022event,
  title={Event-aided direct sparse odometry},
  author={Hidalgo-Carri{\'o}, Javier and Gallego, Guillermo and Scaramuzza, Davide},
  booktitle={Proceedings of the IEEE/CVF Conference on Computer Vision and Pattern Recognition},
  pages={5781--5790},
  year={2022}
}

@article{fraundorfer2012visual,
  title={Visual odometry: Part ii: Matching, robustness, optimization, and applications},
  author={Fraundorfer, Friedrich and Scaramuzza, Davide},
  journal={IEEE Robotics \& Automation Magazine},
  volume={19},
  number={2},
  pages={78--90},
  year={2012},
  publisher={IEEE}
}

@article{engel2017direct,
  title={Direct sparse odometry},
  author={Engel, Jakob and Koltun, Vladlen and Cremers, Daniel},
  journal={IEEE transactions on pattern analysis and machine intelligence},
  volume={40},
  number={3},
  pages={611--625},
  year={2017},
  publisher={IEEE}
}

@article{burri2016euroc,
  title={The EuRoC micro aerial vehicle datasets},
  author={Burri, Michael and Nikolic, Janosch and Gohl, Pascal and Schneider, Thomas and Rehder, Joern and Omari, Sammy and Achtelik, Markus W and Siegwart, Roland},
  journal={The International Journal of Robotics Research},
  volume={35},
  number={10},
  pages={1157--1163},
  year={2016},
  publisher={SAGE Publications Sage UK: London, England}
}

@inproceedings{sturm2012benchmark,
  title={A benchmark for the evaluation of RGB-D SLAM systems},
  author={Sturm, J{\"u}rgen and Engelhard, Nikolas and Endres, Felix and Burgard, Wolfram and Cremers, Daniel},
  booktitle={2012 IEEE/RSJ international conference on intelligent robots and systems},
  pages={573--580},
  year={2012},
  organization={IEEE}
}

@article{yuan2022robust,
  title={Robust Visual Odometry Leveraging Mixture of Manhattan Frames in Indoor Environments},
  author={Yuan, Huayu and Wu, Chengfeng and Deng, Zhongliang and Yin, Jiahui},
  journal={Sensors},
  volume={22},
  number={22},
  pages={8644},
  year={2022},
  publisher={MDPI}
}

@inproceedings{handa2014benchmark,
  title={A benchmark for RGB-D visual odometry, 3D reconstruction and SLAM},
  author={Handa, Ankur and Whelan, Thomas and McDonald, John and Davison, Andrew J},
  booktitle={2014 IEEE international conference on Robotics and automation (ICRA)},
  pages={1524--1531},
  year={2014},
  organization={IEEE}
}

@inproceedings{forster2014svo,
  title={SVO: Fast semi-direct monocular visual odometry},
  author={Forster, Christian and Pizzoli, Matia and Scaramuzza, Davide},
  booktitle={2014 IEEE international conference on robotics and automation (ICRA)},
  pages={15--22},
  year={2014},
  organization={IEEE}
}
\end{document}